\documentclass[journal]{IEEEtran}

\usepackage{cite}
\usepackage{amsmath,amssymb,amsfonts}
\usepackage{graphicx}
\usepackage[table]{xcolor}
\usepackage{booktabs}
\usepackage{diagbox}
\usepackage{multirow}
\usepackage{algorithm}
\usepackage{algpseudocode}
\usepackage{tabularray}
\usepackage{textcomp}
\usepackage{comment}
\usepackage{hyperref}

\definecolor{sectionblue}{RGB}{230,238,246}
\definecolor{DeepRed}{RGB}{160,32,32}
\definecolor{DeepBlue}{RGB}{35,72,140}

\AtBeginDocument{
    \setlength{\abovedisplayskip}{4pt}
    \setlength{\belowdisplayskip}{4pt}
    \setlength{\abovedisplayshortskip}{2pt}
    \setlength{\belowdisplayshortskip}{2pt}
}

\newcommand{\improve}[1]{\textcolor{DeepRed}{#1}}
\newcommand{\degrade}[1]{\textcolor{DeepBlue}{#1}}

\def\BibTeX{{\rm B\kern-.05em{\sc i\kern-.025em b}\kern-.08em
    T\kern-.1667em\lower.7ex\hbox{E}\kern-.125emX}}

\begin{document}

\title{Generative Drifting for Conditional Medical Image Generation}

\author{Zirong Li, Siyuan Mei, Weiwen Wu, Andreas Maier, Lina G\"olz, and Yan Xia
\thanks{Z. Li, L. Gölz, and Y. Xia are with the Department of Orthodontics and Orofacial Orthopedics, Friedrich-Alexander-University Erlangen-Nuremberg, Erlangen, Germany.}
\thanks{Z. Li, and Y. Xia are also with the Department Artificial Intelligence in Biomedical Engineering, Friedrich-Alexander-University Erlangen-Nuremberg, Erlangen, Germany.}
\thanks{S. Mei and A. Maier are with Pattern Recognition Lab, Friedrich-Alexander-University Erlangen-Nuremberg, Erlangen, Germany.}
\thanks{W. Wu is with the Department of Biomedical Engineering, Sun-Yat-sen University, Shenzhen, China.}
\thanks{The corresponding author is Y. Xia. e-mail:yan.xia@fau.de}
}

\maketitle

\begin{abstract}
Conditional medical image generation plays an important role in many clinically relevant imaging tasks. However, existing methods still face a fundamental challenge in balancing inference efficiency, patient-specific fidelity, and distribution-level plausibility, particularly in high-dimensional 3D medical imaging. In this work, we propose GDM, a generative drifting framework that reformulates deterministic medical image prediction as a multi-objective learning problem to jointly promote distribution-level plausibility and patient-specific fidelity while retaining one-step inference. GDM extends drifting to 3D medical imaging through an attractive--repulsive drift that minimizes the discrepancy between the generator pushforward and the target distribution. To enable stable drifting-based learning in 3D volumetric data, GDM constructs a multi-level feature bank from a medical foundation encoder to support reliable affinity estimation and drifting field computation across complementary global, local, and spatial representations. In addition, a gradient coordination strategy in the shared output space improves optimization balance under competing distribution-level and fidelity-oriented objectives. We evaluate the proposed framework on two representative tasks, MRI-to-CT synthesis and sparse-view CT reconstruction. Experimental results show that GDM consistently outperforms a wide range of baselines, including GAN-based, flow-matching-based, and SDE-based generative models, as well as supervised regression methods, while improving the balance among anatomical fidelity, quantitative reliability, perceptual realism, and inference efficiency. These findings suggest that GDM provides a practical and effective framework for conditional 3D medical image generation.
\end{abstract}


\section{Introduction}
\IEEEPARstart{M}{edical} image generation has become an important topic in medical image analysis, with applications extending beyond data augmentation and visualization to clinically relevant image-to-image transformation tasks\cite{zhang2024diffboost}\cite{ozbey2023unsupervised}. In particular, conditional medical image generation aims to recover a diagnostically or physically meaningful target image from a given input. Representative examples include magnetic resonance imaging to computed tomography (MRI-to-CT) synthesis for radiotherapy treatment planning and attenuation correction, and sparse-view CT (SVCT) reconstruction for dose reduction and accelerated acquisition\cite{xing2024cross}\cite{yang2025ct}. In these tasks, the correspondence is largely deterministic, as each input is expected to map to a patient-specific target rather than to one plausible sample among many\cite{rassmann2026regression}. The desired output should therefore be anatomically faithful, quantitatively accurate, and perceptually realistic despite modality discrepancies or incomplete measurements\cite{xia2026tomographic}.

\begin{figure}[!t]
\centering
\includegraphics[width=0.98\linewidth]{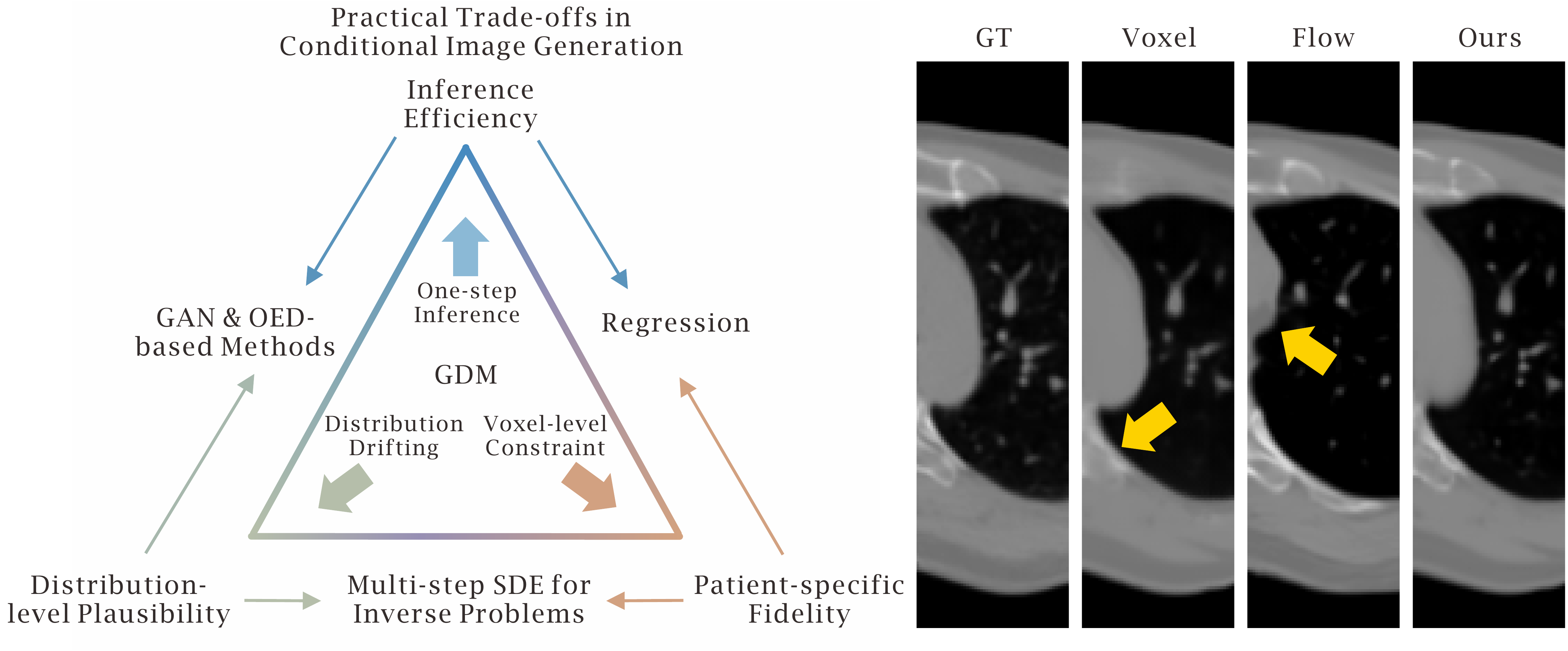}
\caption{Illustration of trade-off among inference efficiency, patient-specific fidelity, and distribution-level plausibility. Regression methods preserve anatomy but may miss fine details, whereas generative methods improve realism with risk of anatomical deviations or hallucinations.}
\label{Fig.Introduction}
\end{figure}

A range of methods have been developed for conditional medical image generation \cite{bahrami2020new,chen2017low, lei2019mri,song2021solving,10577271}. Despite these advances, conditional medical image generation remains fundamentally challenging because clinical tasks impose a three-way tension among inference efficiency, patient-specific fidelity, and distribution-level plausibility. Supervised methods dominated by voxel-wise objectives often behave as deterministic regressors, favoring efficiency and consistency but limiting the recovery of realistic textures and fine details (see Fig.~\ref{Fig.Introduction}, Voxel). In contrast, distribution-aware generative methods can improve perceptual realism, but stronger distribution matching may also introduce subtle anatomical deviations or hallucinated structures that compromise fidelity (see Fig.~\ref{Fig.Introduction}, Flow). In some inverse problems, score-based methods can be combined with data-consistency constraints to better preserve reconstruction trustworthiness\cite{han2024physics}. However, these methods rely on iterative inference, resulting in high inference cost that limits their practicality, especially for volumetric medical imaging\cite{li2024two}.

Recently, drifting models introduced a one-step generative paradigm that performs distribution transport during training rather than iterative refinement at inference \cite{deng2026generative}. It defines an attractive--repulsive drift field that pulls samples toward the data distribution while pushing them away from the current model, and optimizes the generator through a stop-gradient fixed-point regression objective.

This perspective is particularly appealing for medical imaging. In many clinical generation tasks, the desired output is effectively deterministic once the patient-specific input is given, yet the mapping remains ill-posed because of modality mismatch, missing information, or incomplete measurements. A one-step drifting-based model is therefore attractive because it preserves the efficiency and deployment simplicity of direct supervised translation while incorporating an explicit distribution-aware learning signal. Since the drift is imposed during training rather than through repeated stochastic refinement at test time, such a formulation can offer a more favorable compromise between realism and fidelity than pure regression, without inheriting the heavy inference burden of multi-step generative samplers, and is naturally compatible with supervised translation and reconstruction settings.

Nevertheless, extending drifting to conditional medical image generation is nontrivial. A first challenge is reliable affinity estimation for volumetric data. Because drifting is governed by cross-sample affinities, the curse of dimensionality makes such relations unreliable in raw image space. Prior work on generative drifting has shown that stable drifting requires representations that faithfully characterize sample similarity; otherwise, the induced drift field becomes degenerate and distribution optimization becomes unstable or ineffective\cite{deng2026generative}. This difficulty is amplified in medical imaging, where volumetric data are high-dimensional and anatomically heterogeneous: whole-volume representations are often too coarse to capture meaningful local correspondences, while the cost of 3D training limits batch size and weakens the statistical stability of drifting-based estimation. A second challenge is optimization conflict. Conditional medical image generation must jointly satisfy target-domain alignment and patient-specific anatomical fidelity, yet these objectives often induce competing gradients during training. Without effective coordination, optimization can be driven toward anatomically implausible hallucinations or overly smooth solutions that fail to match the target distribution.

Therefore, we propose \textbf{G}enerative \textbf{D}rifting for Conditional \textbf{M}edical Image Generation (GDM), a unified framework that jointly promotes inference efficiency, patient-specific fidelity, and distribution-level plausibility. GDM formulates conditional medical image generation as a coordinated multi-objective problem, where drifting aligns generated outputs with the target distribution while direct supervision preserves patient-specific fidelity. To support stable drift computation in volumetric data, GDM introduces a multi-level feature bank that captures complementary global, local, and spatial information. To reconcile the competing gradients induced by fidelity and distribution alignment, GDM adopts a gradient coordination strategy in the shared output space.

We evaluate GDM on two representative tasks, MRI-to-CT synthesis and SVCT reconstruction. Both require outputs that are anatomically faithful, quantitatively reliable, and perceptually realistic under modality gaps or incomplete measurements, and therefore provide a meaningful evaluation setting for one-step drifting-based generation in medical imaging. The main contributions are summarized as follows:

\begin{itemize}
    \item We propose GDM, a unified generative drifting framework for conditional medical image generation. By reformulating deterministic medical image prediction as a coordinated multi-objective learning problem, GDM provides a practical route toward balancing plausibility, fidelity, and inference efficiency.

    \item To our knowledge, this work is the first to extend drifting to 3D medical imaging by integrating an attractive--repulsive distribution drift into a conditional one-step framework, where it aligns the generator pushforward with the target distribution and serves as a stop-gradient regression signal for training.

    \item We develop a multi-level feature bank derived from a medical foundation model for medically meaningful affinity estimation, and further introduce a shared-output-space gradient coordination strategy to reduce the optimization conflict between distribution-level plausibility and patient-specific fidelity.

    \item We validate GDM on MRI-to-CT synthesis and SVCT reconstruction, demonstrating its effectiveness as a practical one-step generative framework for medical imaging.
\end{itemize}

\section{Related Work}

\subsection{Conditional Medical Image Generation}
Conditional medical image generation tasks such as MRI-to-CT synthesis and sparse-view CT reconstruction are commonly addressed by three paradigms: direct supervised regression\cite{jin2017deep}, adversarial distribution-aware learning\cite{isola2017image}, and generative modeling based on stochastic denoising or continuous transport\cite{xia2026tomographic}.

Direct supervised regression methods learn source-to-target mappings under voxel-wise losses, with representative CNN- and transformer-based models such as FBPConvNet\cite{chen2017low} and SwinUNETR\cite{hatamizadeh2021swin}. They remain widely used because they provide stable optimization and efficient one-step inference\cite{wang2023review}, but their pointwise objective often drives predictions toward conditional averages, yielding blurred interfaces and reduced fine detail\cite{isola2017image}. Adversarial methods alleviate this issue by introducing a discriminator to encourage target-domain realism\cite{isola2017image,lei2019mri,yu2019ea}, producing sharper boundaries and more realistic texture. However, their distribution alignment is implicit and difficult to interpret\cite{yi2019generative}, and training remains sensitive to generator--discriminator balance\cite{mescheder2018training}, especially in 3D settings with limited batch size and strong anatomical constraints.

More recently, diffusion-, score-, and flow-based methods have offered a stronger framework for modeling complex medical image distributions\cite{song2021solving,10577271,lipman2022flow,xia2026tomographic}. Compared with direct regression, they often improve perceptual realism and can incorporate physics- or data-consistency constraints to improve inverse reconstruction reliability\cite{han2024physics}. Yet important limitations remain for deterministic 3D medical imaging: sampling is iterative and costly for volumes\cite{li2024two}, many implementations are 2D or slice-wise and weaken inter-slice consistency\cite{chung2023solving}, and stronger distribution modeling still does not guarantee recovery of the paired patient-specific target rather than merely a plausible target-domain sample. Thus, existing methods still face a trade-off among inference efficiency, patient-specific fidelity, and distribution-level plausibility.

\subsection{Generative Modeling via Drifting}
Among the latest native one-step generative paradigms, drifting shifts distribution transport from inference to optimization\cite{deng2026generative}. Starting from a generator that pushes a simple prior to the model distribution, drifting defines a kernelized attractive--repulsive field: generated samples are pulled toward nearby target samples and pushed away from samples from the current model. The generator is then trained with a stop-gradient fixed-point objective to reproduce its own drifted outputs\cite{hyvarinen1999fast}. Accordingly, distribution alignment is made explicit during training, rather than being delegated to adversarial discrimination or multi-step stochastic refinement, while test-time synthesis remains a single forward pass. This training-time transport perspective is especially attractive for conditional medical image generation, where efficient deployment and anatomy-faithful translation are both essential.

\begin{figure*}[!t]
\centering
\includegraphics[width=0.92\linewidth]{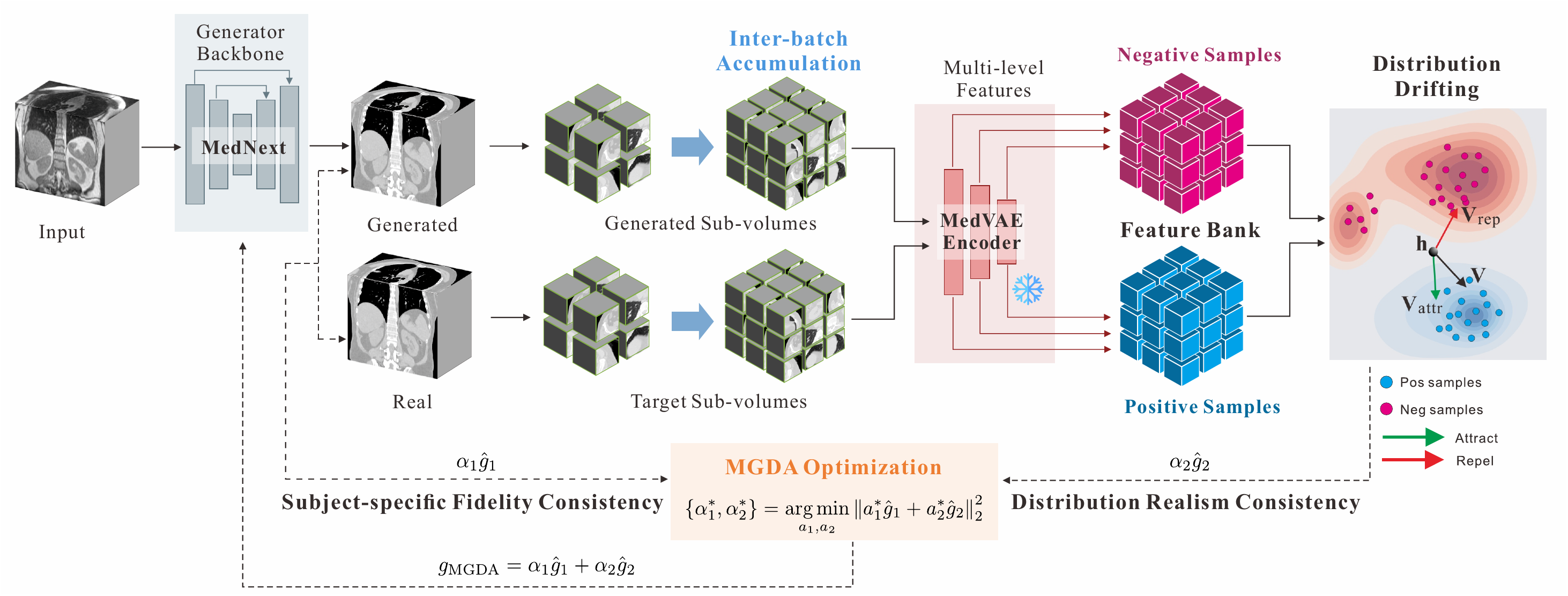}
\caption{Overview of the GDM framework. During training, the generator is jointly optimized by image-domain fidelity supervision and drifting-based distribution alignment, with gradient coordination between the two objectives. The attractive--repulsive distribution drift aligns the generator pushforward with the target distribution by pulling samples toward the target distribution and repelling them from the current model distribution. During inference, only the trained generator is retained to produce the output in one step.}
\label{fig:gdm_pipeline}
\end{figure*}

\begin{figure}[!t]
\centering
\includegraphics[width=0.985\linewidth]{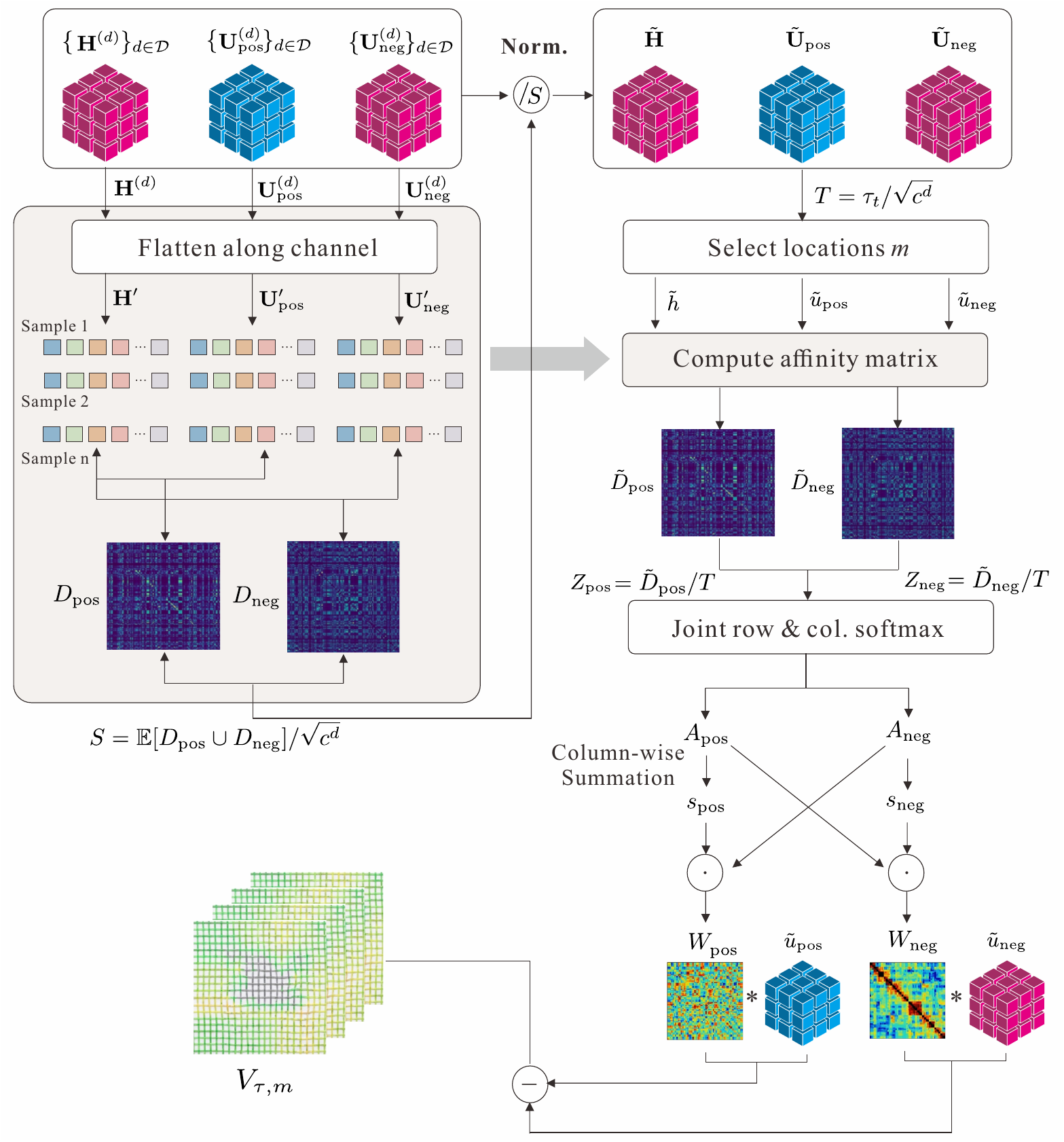}
\caption{Computation of the distribution drifting field. Positive and negative affinities are estimated from target and generated reference features and combined into the final push--pull field.}
\label{fig:compute_V}
\end{figure}

\section{Methodology}
In this section, we present GDM. We first formulate conditional medical image generation as coordinated fidelity--distribution learning. We then introduce drifting distribution alignment in a structured multi-level feature bank, followed by a gradient coordination scheme in the shared output space. An overview of GDM is shown in Fig.~\ref{fig:gdm_pipeline}.

\subsection{Problem Formulation}
The goal is to learn a patient-specific mapping from a source image to its corresponding target image. Let $x$ denote the source image, $y_\mathrm{gt}$ the paired patient-specific target image, and $y = G_{\theta}(x)$ the generated target image, where $G_{\theta}$ is a generator parameterized by $\theta$. For such deterministic tasks, the desired solution should satisfy two requirements simultaneously: 1) strong fidelity to the paired target, and 2) realistic alignment with the target-domain distribution. Accordingly, we formulate the task as a bi-objective optimization problem:
\begin{equation}
\min_{\theta}
\left(
\mathcal{L}_{\mathrm{fidelity}}(\theta),\;
\mathcal{L}_{\mathrm{dist}}(\theta)
\right),
\label{eq:overall_objective}
\end{equation}
where $\mathcal{L}_{\mathrm{fidelity}}$ enforces patient-specific fidelity, and $\mathcal{L}_{\mathrm{dist}}$ promotes alignment with the target distribution.

Under paired supervision, fidelity is directly observable from the target image and can be explicitly enforced. However, the target-domain distribution is not explicitly parameterized or directly supervised; instead, it is only implicitly reflected by the available target samples. As a result, distribution alignment behaves as a more implicit optimization objective during training, making it difficult to optimize in an effective manner. To address this difficulty, GDM introduces \emph{distribution drifting} as an explicit mechanism to progressively drive the generated distribution toward the target distribution. The resulting fidelity and distribution objectives are then coordinated through a gradient coordination strategy.

\subsection{Distribution Drifting}
To explicitly optimize the distribution term in Eq.~(\ref{eq:overall_objective}), GDM introduces \emph{distribution drifting} as a training-time mechanism for distribution alignment. Concretely, this mechanism is realized by a drift field that assigns each generated sample a movement direction toward the target distribution, so that sample-wise evolution progressively reduces the discrepancy between the generated and target distributions and reaches equilibrium when the two are best aligned.

\subsubsection{Training-time Drift Dynamics}
We describe drifting from the perspective of generator outputs during training. At iteration $i$, let $y_i = G_{\theta_i}(x)$ denote the generated target image for source image $x$, and let $q_i$ be the pushforward distribution induced by $G_{\theta_i}$. With respect to the target-domain distribution $p$, we define a drift field $V_{p,q_i}(\cdot)$. A one-step drifting update is
\begin{equation}
y_{i+1}
=
y_i + V_{p,q_i}(y_i).
\label{eq:drift_update}
\end{equation}
This moves the current generator output so that the induced distribution $q_i$ is progressively transported toward $p$.

A meaningful drift field should satisfy equilibrium consistency: once the generated distribution matches the target distribution, no further transport is needed. Formally,
\begin{equation}
q = p
\;\Longrightarrow\;
V_{p,q}(y)=0,
\qquad \forall y.
\label{eq:drift_equilibrium}
\end{equation}
Under this property, the optimal generator is characterized by outputs that are stationary under the drift update.

This yields a fixed-point view of training. At equilibrium, the generator output $y_{\mathrm{eq}} = G_{\theta_{\mathrm{eq}}}(x)$ should satisfy
\begin{equation}
y_{\mathrm{eq}}
=
y_{\mathrm{eq}}
+
V_{p,q_{\mathrm{eq}}}(y_{\mathrm{eq}}),
\label{eq:fixed_point_equilibrium}
\end{equation}
where $q_{\mathrm{eq}}$ is the equilibrium pushforward distribution induced by $G_{\theta_{\mathrm{eq}}}$. Accordingly, training can be viewed as searching for generator parameters whose outputs satisfy this fixed-point condition, motivating the iteration
\begin{equation}
G_{\theta_{i+1}}(x)
\leftarrow
G_{\theta_i}(x) + V_{p,q_i}\bigl(G_{\theta_i}(x)\bigr).
\label{eq:fixed_point_iteration}
\end{equation}

In practice, GDM does not optimize Eq.~(\ref{eq:fixed_point_iteration}) by back-propagating through the drift field, because $V_{p,q_i}$ depends on the evolving pushforward distribution $q_i$. Instead, one drifting step is realized as a stop-gradient regression target:
\begin{equation}
\mathcal{L}_{\mathrm{drift}}
=
\mathbb{E}_{x}
\left[
\left\|
G_{\theta_i}(x)
-
\operatorname{sg}
\Bigl(
G_{\theta_i}(x) + V_{p,q_i}\bigl(G_{\theta_i}(x)\bigr)
\Bigr)
\right\|_2^2
\right],
\label{eq:drift_surrogate_loss}
\end{equation}
where $\operatorname{sg}(\cdot)$ denotes the stop-gradient operator.

For Eq.~(\ref{eq:drift_surrogate_loss}) to be useful, the drift field must capture meaningful distribution alignment through affinity estimation over target and generated samples. However, such estimation is unreliable in raw voxel space for volumetric medical data because of high dimensionality and the limited ability of whole-volume representations to capture medically meaningful local correspondences. We therefore instantiate drifting over a collection of complementary feature spaces indexed by $d\in\mathcal{D}$, where each $d$ denotes a descriptor family defined on local volumetric representations. The continuous and discrete realizations of the drift field are described next, and the construction of $\mathcal{D}$ is given later in the Section \ref{sec:mfbc}.

\subsubsection{Continuous Distribution Drift Field}

We next instantiate the abstract drift field in Eq.~(\ref{eq:drift_update}) within each feature space $d\in\mathcal{D}$. Let $\phi^{(d)}$ denote the feature extractor associated with the $d$-th feature space. For a generated image $y$, we denote its feature representation by $h^{(d)} = \phi^{(d)}(y)$. Correspondingly, let $q_\theta^{(d)} := (\phi^{(d)})_\# q_\theta$ and $p^{(d)} := (\phi^{(d)})_\# p$ denote the generated and target feature distributions in the $d$-th feature space, respectively.
For a generated feature $h^{(d)} \sim q_\theta^{(d)}$, we instantiate the drift field as
\begin{equation}
V(h^{(d)})
=
V_{\mathrm{attr}}(h^{(d)})
-
V_{\mathrm{rep}}(h^{(d)}).
\label{eq:drift_field}
\end{equation}
The attraction component is defined by
\begin{equation}
V_{\mathrm{attr}}(h^{(d)})
=
\frac{
\mathbb{E}_{u^+ \sim p^{(d)}}
\left[
k(h^{(d)},u^+)\bigl(u^+ - h^{(d)}\bigr)
\right]
}{
\mathbb{E}_{u^+ \sim p^{(d)}}
\left[
k(h^{(d)},u^+)
\right]
},
\label{eq:attraction}
\end{equation}
and the repulsion component is defined by
\begin{equation}
V_{\mathrm{rep}}(h^{(d)})
=
\frac{
\mathbb{E}_{u^- \sim q_\theta^{(d)}}
\left[
k(h^{(d)},u^-)\bigl(u^- - h^{(d)}\bigr)
\right]
}{
\mathbb{E}_{u^- \sim q_\theta^{(d)}}
\left[
k(h^{(d)},u^-)
\right]
},
\label{eq:repulsion}
\end{equation}
where $k(\cdot,\cdot)$ denotes a similarity function. In practice, we use Laplacian-style distance logits of the form $-\lVert y-x\rVert/\epsilon$, followed by bidirectional softmax normalization to obtain the final affinities. Under this formulation, $V_{\mathrm{attr}}(h^{(d)})$ computes a normalized kernel-weighted mean displacement from $h^{(d)}$ toward the target feature distribution, whereas $V_{\mathrm{rep}}(h^{(d)})$ computes the corresponding displacement toward the current generated feature distribution. Their difference therefore defines a feature-space drift direction that encourages alignment with the target distribution while discouraging concentration in regions already occupied by generated features.

\subsubsection{Drifting Field Computation}

The continuous drift field in Eqs.~(\ref{eq:attraction})--(\ref{eq:repulsion}) is defined as expectations over feature distributions and is approximated in practice by empirical averages over finite feature sets, yielding the discrete empirical drift computation shown in Fig.~\ref{fig:compute_V}.

For drift computation, target-domain reference samples are treated as positive samples, whereas generated samples drawn from the current pushforward distribution are treated as negative samples. Rather than computing drift on full volumes, we evaluate it on batch-level features over the feature-space collection $\mathcal{D}$. For a given feature space $d$, let
\begin{equation}
\mathbf{H}^{(d)},\;
\mathbf{U}_{\mathrm{pos}}^{(d)},\;
\mathbf{U}_{\mathrm{neg}}^{(d)}
\in
\mathbb{R}^{N \times M_d \times C_d},
\label{eq:feature_tensor}
\end{equation}
denote the synthesized feature tensor, positive target feature tensor, and negative generated feature tensor, respectively, where $N$ is the batch size, $M_d$ is the number of sampled sub-volumes, spatial locations, or pooled neighborhood elements, and $C_d$ is the channel dimension. Here, $\mathbf{H}^{(d)}$ is obtained by applying $\phi^{(d)}$ to the generated image $y = G_\theta(x)$, $\mathbf{U}_{\mathrm{pos}}^{(d)}$ from the paired target image $y_\mathrm{gt}$, and $\mathbf{U}_{\mathrm{neg}}^{(d)}$ from generated reference samples drawn from the current generated distribution $q_\theta$. Thus, negative samples provide a finite empirical approximation to $q_\theta^{(d)}$, while trivial self-matches are suppressed by a masking constant $\mu_{\mathrm{mask}} > 0$.

To improve comparability across feature spaces, we first compute global positive and negative distance matrices from flattened empirical feature sets:
\begin{equation}
\begin{aligned}
D_{\mathrm{pos}}^{(d)}
&=
\operatorname{cdist}\!\left(\mathbf{H}'^{(d)}, \mathbf{U}_{\mathrm{pos}}'^{(d)}\right),\\
D_{\mathrm{neg}}^{(d)}
&=
\operatorname{cdist}\!\left(\mathbf{H}'^{(d)}, \mathbf{U}_{\mathrm{neg}}'^{(d)}\right)
+
\mu_{\mathrm{mask}} I.
\end{aligned}
\label{eq:global_distances}
\end{equation}
where the primed tensors denote flattened versions of $\mathbf{H}^{(d)}$, $\mathbf{U}_{\mathrm{pos}}^{(d)}$, and $\mathbf{U}_{\mathrm{neg}}^{(d)}$. We then estimate a global distance scale
\begin{equation}
S^{(d)} = \mathbb{E}\!\left[D_{\mathrm{pos}}^{(d)} \cup D_{\mathrm{neg}}^{(d)}\right] / \sqrt{C_d},
\label{eq:global_scale}
\end{equation}
and normalize the feature tensors as
\begin{equation}
\tilde{\mathbf{H}}^{(d)}
=
\frac{\mathbf{H}^{(d)}}{S^{(d)}},
\qquad
\tilde{\mathbf{U}}_{\mathrm{pos}}^{(d)}
=
\frac{\mathbf{U}_{\mathrm{pos}}^{(d)}}{S^{(d)}},
\qquad
\tilde{\mathbf{U}}_{\mathrm{neg}}^{(d)}
=
\frac{\mathbf{U}_{\mathrm{neg}}^{(d)}}{S^{(d)}}.
\label{eq:feature_normalization}
\end{equation}
When $\mu_{\mathrm{mask}} I$ is also included in the global scale estimation, the resulting scale is slightly inflated, especially for feature branches with small raw distances. This compresses the dynamic range of normalized distances, yields flatter affinities, and therefore leads to more generative behavior.

For each local feature location $m \in \{1,\dots,M_d\}$ and temperature $\tau \in \Lambda$, we construct a joint affinity matrix
\begin{equation}
A_m^{(d,\tau)}
=
\sqrt{
\operatorname{softmax}_{\mathrm{r}}\!\left(Z_m^{(d,\tau)}\right)
\odot
\operatorname{softmax}_{\mathrm{c}}\!\left(Z_m^{(d,\tau)}\right)
},
\label{eq:joint_affinity}
\end{equation}
where $Z_m^{(d,\tau)}$ is formed by concatenating the temperature-scaled positive and negative local distance matrices, and $\operatorname{softmax}_{\mathrm{r}}(\cdot)$ and $\operatorname{softmax}_{\mathrm{c}}(\cdot)$ denote softmax normalization along the candidate and query dimensions, respectively. Splitting $A_m^{(d,\tau)}$ into positive and negative parts gives
\begin{equation}
\left(
A_{\mathrm{pos},m}^{(d,\tau)},
A_{\mathrm{neg},m}^{(d,\tau)}
\right)
=
\operatorname{split}\!\left(A_m^{(d,\tau)}, [N,N], \mathrm{dim}=1\right),
\label{eq:split_affinity}
\end{equation}
from which we define aggregated responses
\begin{equation}
s_{\mathrm{pos},m}^{(d,\tau)}=\sum_j [A_{\mathrm{pos},m}^{(d,\tau)}]_{:,j},\quad
s_{\mathrm{neg},m}^{(d,\tau)}=\sum_j [A_{\mathrm{neg},m}^{(d,\tau)}]_{:,j}.
\label{eq:aggregated_response}
\end{equation}
and cross push--pull weights
\begin{equation}
\begin{aligned}
W_{\mathrm{pos},m}^{(d,\tau)}
&=
A_{\mathrm{pos},m}^{(d,\tau)} \odot s_{\mathrm{neg},m}^{(d,\tau)}
\hspace{1.2em}
W_{\mathrm{neg},m}^{(d,\tau)}
=
A_{\mathrm{neg},m}^{(d,\tau)} \odot s_{\mathrm{pos},m}^{(d,\tau)}.
\end{aligned}
\label{eq:cross_weight}
\end{equation}

The resulting empirical local drift estimate is
\begin{equation}
V_{\tau,m}^{(d)}
=
W_{\mathrm{pos},m}^{(d,\tau)}\tilde{u}_{\mathrm{pos},m}
-
W_{\mathrm{neg},m}^{(d,\tau)}\tilde{u}_{\mathrm{neg},m},
\label{eq:local_drift}
\end{equation}
which is then aggregated over all locations and temperatures to obtain the final empirical drift field:
\begin{equation}
V^{(d)} = \sum_{\tau \in \Lambda} V_{\tau}^{(d)} / \left( \|V_{\tau}^{(d)}\|_F / \sqrt{N M_d C_d} + \varepsilon \right),
\label{eq:drift_aggregation}
\end{equation}
where $V_{\tau}^{(d)}$ denotes the temperature-specific field obtained by stacking the local drift estimates over locations. The overall pipeline is described in Algorithm. \ref{alg:compute_V}.

\subsubsection{Training Objective for Drifting}

Given the empirical drifting field $V^{(d)}$, we realize the drift update through a stop-gradient regression objective. The drifted target for the current feature tensor is $\tilde{\mathbf{H}}^{(d)} + V^{(d)}$. However, directly regressing to this evolving target would allow the network to reshape the feature geometry in a way that weakens the drift constraint itself. To avoid this behavior, we use the stop-gradient operator $\operatorname{sg}(\cdot)$ to define a detached target:
\begin{equation}
\mathbf{H}^{(d)}_{\mathrm{tgt}}
=
\operatorname{sg}
\left(
\tilde{\mathbf{H}}^{(d)} + V^{(d)}
\right).
\label{eq:anchor}
\end{equation}

The distribution alignment term is then instantiated as a distribution drifting loss:
\begin{equation}
\mathcal{L}_{\mathrm{dist}}(\theta)
\approx
\mathcal{L}_{\mathrm{drift}}(\theta)
=
\sum_{d \in \mathcal{D}}
\frac{1}{2}
\left\|
\tilde{\mathbf{H}}^{(d)} - \mathbf{H}^{(d)}_{\mathrm{tgt}}
\right\|_2^2.
\label{eq:drift_loss}
\end{equation}

Since $\mathbf{H}^{(d)}_{\mathrm{tgt}}$ is detached from backpropagation, the gradient of the drift loss with respect to the current feature tensor is
\begin{equation}
\nabla_{\tilde{\mathbf{H}}^{(d)}} \mathcal{L}_{\mathrm{drift}}
=
\tilde{\mathbf{H}}^{(d)}
-
\operatorname{sg}
\left(
\tilde{\mathbf{H}}^{(d)} + V^{(d)}
\right)
=
- V^{(d)}.
\label{eq:drift_gradient}
\end{equation}
Therefore, the designed drifting field is incorporated into training through a stop-gradient regression objective. This objective provides a practical realization of the fixed-point drift update in Eq.~(\ref{eq:fixed_point_iteration}). The concrete construction of the feature-space collection $\mathcal{D}$ is described below.

\begin{algorithm}[!t]
\caption{Compute Distribution Drifting Field}
\label{alg:compute_V}
\small
\setlength{\baselineskip}{1.2\baselineskip}
\renewcommand{\arraystretch}{1}
\begin{algorithmic}[1]
\Require $\{(\mathbf{H}^{(d)}, \mathbf{U}_{\mathrm{pos}}^{(d)}, \mathbf{U}_{\mathrm{neg}}^{(d)})\}_{d\in\mathcal D}$,
$\mathbf{H}^{(d)}, \mathbf{U}_{\mathrm{pos}}^{(d)}, \mathbf{U}_{\mathrm{neg}}^{(d)} \in \mathbb{R}^{N \times M_d \times C_d}$ $\Lambda=\{\tau_1,\dots,\tau_K\}$, identity matrix $I$, masking constant $\mu_{\mathrm{mask}}$, small constant $\varepsilon$

\ForAll{$d \in \mathcal D$}
    \State $\mathbf{H}' \gets \operatorname{reshape}_{[-1,C_d]}(\mathbf{H}^{(d)})$
    \State $\mathbf{U}'_{\mathrm{pos}} \gets \operatorname{reshape}_{[-1,C_d]}(\mathbf{U}_{\mathrm{pos}}^{(d)})$
    \State $\mathbf{U}'_{\mathrm{neg}} \gets \operatorname{reshape}_{[-1,C_d]}(\mathbf{U}_{\mathrm{neg}}^{(d)})$
    \State $D_{\mathrm{pos}} \gets \operatorname{cdist}(\mathbf{H}', \mathbf{U}'_{\mathrm{pos}})$
    \State $D_{\mathrm{neg}} \gets \operatorname{cdist}(\mathbf{H}', \mathbf{U}'_{\mathrm{neg}}) + \mu_{\mathrm{mask}} I$
    \State $S \gets \mathbb{E}[D_{\mathrm{pos}} \cup D_{\mathrm{neg}}] / \sqrt{C_d}$

    \State $\tilde{\mathbf{H}} \gets \mathbf{H}^{(d)} / S$
    \State $\tilde{\mathbf{U}}_{\mathrm{pos}} \gets \mathbf{U}_{\mathrm{pos}}^{(d)} / S$
    \State $\tilde{\mathbf{U}}_{\mathrm{neg}} \gets \mathbf{U}_{\mathrm{neg}}^{(d)} / S$
    \State $V^{(d)} \gets 0$

    \ForAll{$\tau \in \Lambda$}
        \State $T \gets \tau \sqrt{C_d}$

        \For{$m = 1,\dots,M_d$}
            \State $\tilde{h} \gets (\tilde{\mathbf{H}})_{:,m,:}$
            \State $\tilde{u}_{\mathrm{pos}} \gets (\tilde{\mathbf{U}}_{\mathrm{pos}})_{:,m,:}$
            \State $\tilde{u}_{\mathrm{neg}} \gets (\tilde{\mathbf{U}}_{\mathrm{neg}})_{:,m,:}$

            \State $\tilde{D}_{\mathrm{pos}} \gets \operatorname{cdist}(\tilde{h}, \tilde{u}_{\mathrm{pos}})$
            \State $\tilde{D}_{\mathrm{neg}} \gets \operatorname{cdist}(\tilde{h}, \tilde{u}_{\mathrm{neg}}) + \mu_{\mathrm{mask}} I$

            \State $Z_{\mathrm{pos}} \gets -\tilde{D}_{\mathrm{pos}} / T$
            \State $Z_{\mathrm{neg}} \gets -\tilde{D}_{\mathrm{neg}} / T$
            \State $Z \gets \operatorname{concat}(Z_{\mathrm{pos}}, Z_{\mathrm{neg}}, \mathrm{dim}=1)$
            \State $A \gets \sqrt{
            \operatorname{softmax}_{\mathrm{r}}(Z)
            \odot
            \operatorname{softmax}_{\mathrm{c}}(Z)}$

            \State $(A_{\mathrm{pos}}, A_{\mathrm{neg}}) \gets \operatorname{split}(A, [N,N], \mathrm{dim}=1)$
            \State $s_{\mathrm{pos}} \gets \sum_j [A_{\mathrm{pos}}]_{:,j}$
            \State $s_{\mathrm{neg}} \gets \sum_j [A_{\mathrm{neg}}]_{:,j}$
            \State $W_{\mathrm{pos}} \gets A_{\mathrm{pos}} \odot s_{\mathrm{neg}}$
            \State $W_{\mathrm{neg}} \gets A_{\mathrm{neg}} \odot s_{\mathrm{pos}}$
            \State $V_{\tau,m} \gets W_{\mathrm{pos}} \tilde{u}_{\mathrm{pos}} - W_{\mathrm{neg}} \tilde{u}_{\mathrm{neg}}$
        \EndFor

        \State $V_{\tau} \gets \operatorname{stack}(\{V_{\tau,m}\}_{m=1}^{M_d}, \mathrm{dim}=1)$
        \State $V^{(d)} \gets V^{(d)} + V_{\tau} / \left( \|V_{\tau}\|_F / \sqrt{N M_d C_d} + \varepsilon \right)$
    \EndFor
\EndFor

\State \Return $\{V^{(d)}\}_{d\in\mathcal D}$
\end{algorithmic}
\end{algorithm}

\begin{figure}[!t]
\centering
\includegraphics[width=0.98\linewidth]{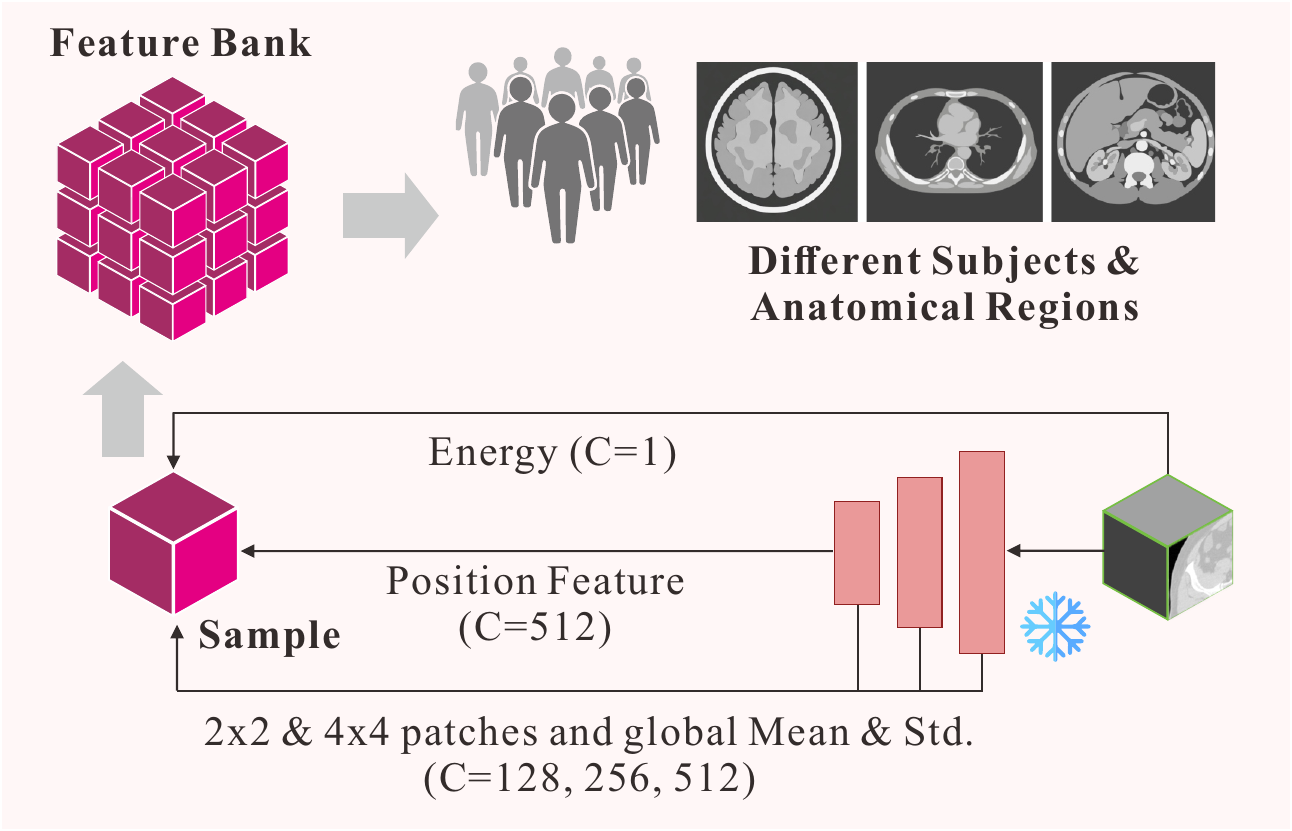}
\caption{Illustration of the multi-level feature bank used for drifting. Local sub-volumes are encoded by a separately fine-tuned and frozen MedVAE-3D encoder to form complementary descriptor families that support affinity estimation in structured feature spaces.}
\label{fig:feature_bank}
\end{figure}

\subsubsection{Multi-level Feature Bank Construction}
\label{sec:mfbc}
To support reliable affinity estimation for drifting, as shown in Fig.~\ref{fig:feature_bank}, we construct a multi-level feature bank based on two complementary principles.

\noindent\textbf{a) Sample-level semantic coverage.}
Reliable drifting requires a semantically rich empirical bank spanning different patients and anatomical structures. Whole-volume representations are often too coarse, whereas local substructures exhibit stronger cross-subject correspondence. We therefore decompose each 3D volume into local sub-volumes to enrich the empirical support for drift estimation, improve the statistical stability of distribution modeling, and reduce the computational burden of volumetric processing. In practice, sub-volumes are randomly sampled from the current mini-batch and can be aggregated across devices under distributed training.

\noindent\textbf{b) Feature-level semantic complementarity.}
Reliable affinity estimation also requires representations that faithfully reflect sample similarity. We therefore adopt MedVAE-3D\cite{varma2025medvae}, a medical foundation encoder pretrained on more than 30,000 multimodal 3D medical images, as the drifting backbone. By mapping sub-volumes from raw voxel space into lower-dimensional structured feature spaces, MedVAE-3D alleviates the curse of dimensionality in volumetric medical data and makes affinity estimation more tractable. Its hierarchical and anatomically meaningful representations further provide a stable basis for constructing complementary descriptor families.

Specifically, we use a separately fine-tuned and frozen MedVAE-3D encoder $F_{\phi}$ as the drifting feature extractor, and define the feature-space collection as
\begin{equation}
\mathcal{D}
=
\mathcal{D}_{\mathrm{energy}}
\cup
\mathcal{D}_{\mathrm{global}}
\cup
\mathcal{D}_{\mathrm{local}}
\cup
\mathcal{D}_{\mathrm{spatial}},
\label{eq:feature_bank}
\end{equation}
where $\mathcal{D}_{\mathrm{energy}}$ contains channel-wise energy descriptors from the encoder input, $\mathcal{D}_{\mathrm{global}}$ contains global mean and standard-deviation descriptors of feature maps, $\mathcal{D}_{\mathrm{local}}$ contains semantically rich local descriptors from the deepest encoder stage, and $\mathcal{D}_{\mathrm{spatial}}$ contains multi-scale neighborhood statistics obtained by pooling feature maps over local windows (e.g., $2\times2\times2$ and $4\times4\times4$). Extracted from multiple selected stages of the separately fine-tuned and frozen MedVAE-3D encoder, these descriptors capture complementary information from global response statistics to local anatomy and spatial organization, thereby supporting stable affinity estimation.

\subsection{Gradient Coordination for Fidelity and Distribution}
Once the fidelity and distribution objectives are defined, the key challenge becomes joint optimization, because the two objectives may induce conflicting descent directions, making a fixed linear combination unreliable\cite{liu2021conflict}. Excessive fidelity weighting drives the model toward deterministic regression, whereas excessive drifting weighting can damage patient-specific correspondence. To address this issue, GDM adopts an MGDA-based gradient coordination strategy in the shared output space\cite{sener2018multi}.

Let $y = G_{\theta}(x)$ denote the generated output. In the main setting, we consider two objectives:
\begin{equation}
L_1 = \mathcal{L}_{\mathrm{fidelity}},
\qquad
L_2 = \lambda \mathcal{L}_{\mathrm{drift}},
\label{eq:mgda_objectives}
\end{equation}
where $\lambda$ is the drift weighting coefficient applied before MGDA. We then compute their gradients with respect to the shared representation $y$:
\begin{equation}
\mathbf{g}_i = \nabla_y L_i, \qquad i\in\{1,2\}.
\label{eq:mgda_output_grad}
\end{equation}

From these gradients, we construct the Gram matrix
\begin{equation}
H_{ij} = \mathbf{g}_i^\top \mathbf{g}_j,
\label{eq:mgda_gram}
\end{equation}
and solve for convex combination coefficients on the probability simplex:
\begin{equation}
\boldsymbol{\alpha}^{\ast}
=
\arg\min_{\boldsymbol{\alpha}}
\boldsymbol{\alpha}^{\top} H \boldsymbol{\alpha}
\quad
\text{s.t.}
\quad
\alpha_i \ge 0,\;
\sum_i \alpha_i = 1.
\label{eq:mgda_alpha}
\end{equation}
In practice, Eq.~(\ref{eq:mgda_alpha}) is solved numerically by projected gradient descent on the simplex.

The resulting coefficients define the training objective
\begin{equation}
\mathcal{L}_{\mathrm{total}}
=
\sum_i \alpha_i^{\ast} L_i
=
\alpha_1^{\ast}\mathcal{L}_{\mathrm{fidelity}}
+
\alpha_2^{\ast}\lambda\mathcal{L}_{\mathrm{drift}},
\label{eq:mgda_total_loss}
\end{equation}
which is then back-propagated through the generator parameters. In this way, patient-specific fidelity and distribution alignment are adaptively balanced according to the geometry of their shared-output gradients.

\begin{figure*}[!t]
\centering
\includegraphics[width=0.985\linewidth]{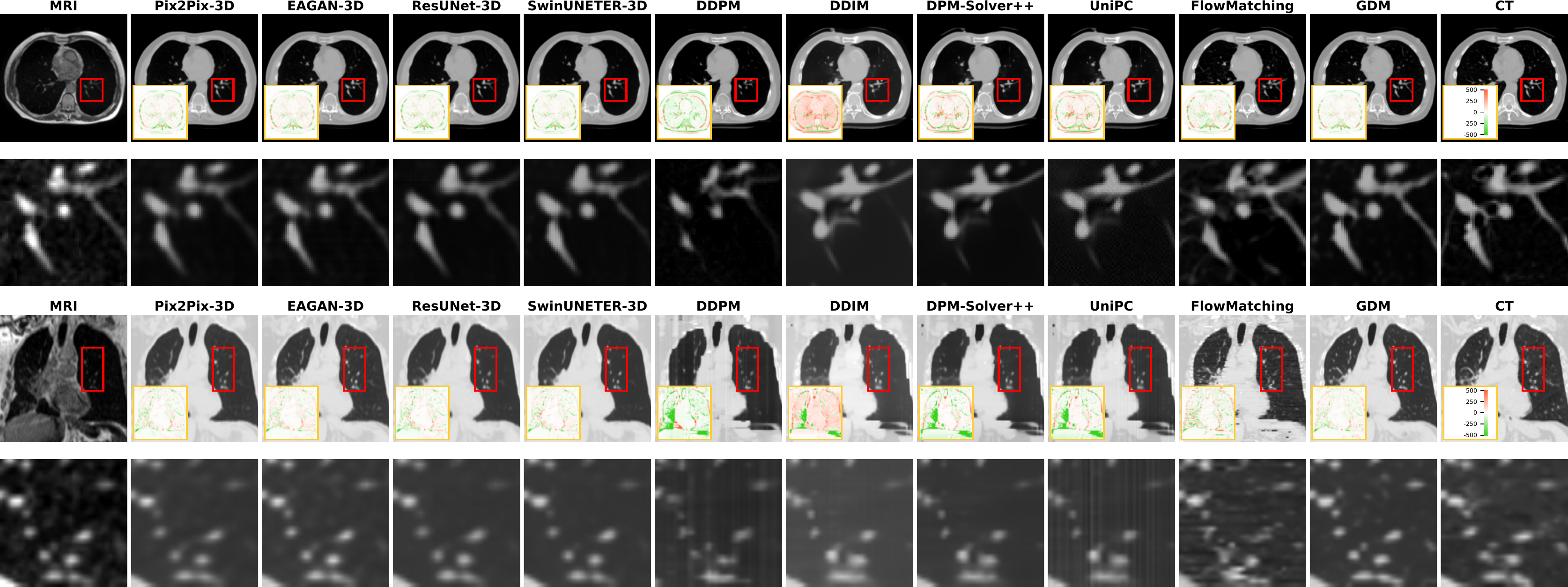}
\caption{MRI-to-CT synthesis results. Rows 1--2 show an axial case (WW/WL = 1500/-200 HU), and rows 3--4 show a coronal case (WW/WL = 1400/-500 HU). Rows 2 and 4 show enlarged ROIs corresponding to the red boxes. Insets are signed error maps with respect to the reference CT.}
\label{fig:mri_ct_vis}
\end{figure*}

\begin{table*}[!t]
\centering
\caption{Quantitative evaluation of jointly trained MRI-to-CT synthesis, reported separately on Head-and-Neck, Thoracic, and Abdominal scans from the SynthRAD2025 dataset. Best results are highlighted in bold.}
\label{tab:mri_ct_results}
\renewcommand{\arraystretch}{1.5}
\setlength{\tabcolsep}{3pt}
\resizebox{0.98\textwidth}{!}{%
\begin{tabular}{lcccccccccccc}
\toprule
\rowcolor{sectionblue}
\multicolumn{13}{c}{\textbf{MRI-to-CT Synthesis Results on SynthRAD2025 Dataset}\cite{thummerer2025synthrad2025}} \\
\midrule
\multirow{2}{*}{\textbf{Method}}
& \multicolumn{4}{c}{\textbf{Head-and-Neck}}
& \multicolumn{4}{c}{\textbf{Thoracic}}
& \multicolumn{4}{c}{\textbf{Abdominal}} \\
\cmidrule(lr){2-5}\cmidrule(lr){6-9}\cmidrule(lr){10-13}
& \textbf{MAE}$\downarrow$ & \textbf{MS-SSIM}$\uparrow$ & \textbf{Dice}$\uparrow$ & \textbf{HD95}$\downarrow$
& \textbf{MAE}$\downarrow$ & \textbf{MS-SSIM}$\uparrow$ & \textbf{Dice}$\uparrow$ & \textbf{HD95}$\downarrow$
& \textbf{MAE}$\downarrow$ & \textbf{MS-SSIM}$\uparrow$ & \textbf{Dice}$\uparrow$ & \textbf{HD95}$\downarrow$ \\
\midrule
Pix2Pix-3D\cite{isola2017image}
& 80.25 $\pm$ 19.60 & 0.947 $\pm$ 0.030 & 0.777 $\pm$ 0.076 & 4.616 $\pm$ 1.486
& 68.83 $\pm$ 23.46 & 0.920 $\pm$ 0.053 & 0.543 $\pm$ 0.107 & 21.17 $\pm$ 9.015
& 70.30 $\pm$ 13.98 & 0.876 $\pm$ 0.028 & 0.443 $\pm$ 0.090 & 34.14 $\pm$ 12.11 \\

EAGAN-3D\cite{yu2019ea}
& 79.04 $\pm$ 19.00 & 0.950 $\pm$ 0.027 & 0.792 $\pm$ 0.087 & 4.271 $\pm$ 1.315
& 67.27 $\pm$ 23.51 & 0.924 $\pm$ 0.054 & 0.612 $\pm$ 0.098 & 12.84 $\pm$ 5.341
& 68.98 $\pm$ 13.56 & 0.883 $\pm$ 0.026 & 0.543 $\pm$ 0.083 & 19.99 $\pm$ 5.694 \\

ResUNet-3D\cite{xiao2018weighted}
& 83.69 $\pm$ 19.17 & 0.945 $\pm$ 0.027 & 0.764 $\pm$ 0.088 & 5.622 $\pm$ 3.072
& 72.59 $\pm$ 23.71 & 0.915 $\pm$ 0.052 & 0.502 $\pm$ 0.105 & 24.15 $\pm$ 9.192
& 73.30 $\pm$ 16.74 & 0.869 $\pm$ 0.037 & 0.430 $\pm$ 0.085 & 36.95 $\pm$ 8.938 \\

SwinUNETR-3D\cite{hatamizadeh2021swin}
& 81.42 $\pm$ 19.66 & 0.944 $\pm$ 0.030 & 0.764 $\pm$ 0.090 & 5.498 $\pm$ 2.944
& 69.35 $\pm$ 23.21 & 0.917 $\pm$ 0.053 & 0.517 $\pm$ 0.097 & 21.52 $\pm$ 8.221
& 71.93 $\pm$ 15.87 & 0.870 $\pm$ 0.034 & 0.433 $\pm$ 0.091 & 35.09 $\pm$ 8.451 \\

DDPM\cite{ho2020denoising}
& 148.6 $\pm$ 35.14 & 0.854 $\pm$ 0.050 & 0.670 $\pm$ 0.107 & 6.474 $\pm$ 1.834
& 129.9 $\pm$ 13.12 & 0.800 $\pm$ 0.043 & 0.532 $\pm$ 0.046 & 13.47 $\pm$ 3.603
& 124.5 $\pm$ 8.380 & 0.738 $\pm$ 0.029 & 0.486 $\pm$ 0.060 & 18.32 $\pm$ 3.222 \\

DDIM\cite{song2020denoising}
& 229.5 $\pm$ 43.99 & 0.838 $\pm$ 0.044 & 0.674 $\pm$ 0.088 & 6.857 $\pm$ 1.704
& 177.2 $\pm$ 25.78 & 0.786 $\pm$ 0.044 & 0.512 $\pm$ 0.071 & 14.19 $\pm$ 4.275
& 156.7 $\pm$ 30.59 & 0.761 $\pm$ 0.041 & 0.491 $\pm$ 0.078 & 18.07 $\pm$ 5.706 \\

DPM-Solver++\cite{lu2025dpm}
& 181.0 $\pm$ 38.67 & 0.854 $\pm$ 0.042 & 0.685 $\pm$ 0.086 & 6.461 $\pm$ 1.416
& 145.2 $\pm$ 21.01 & 0.797 $\pm$ 0.052 & 0.508 $\pm$ 0.068 & 14.87 $\pm$ 4.735
& 120.8 $\pm$ 20.35 & 0.762 $\pm$ 0.037 & 0.481 $\pm$ 0.085 & 19.19 $\pm$ 5.591 \\

UniPC\cite{zhao2023unipc}
& 183.6 $\pm$ 38.94 & 0.852 $\pm$ 0.042 & 0.686 $\pm$ 0.086 & 6.450 $\pm$ 1.472
& 147.2 $\pm$ 20.33 & 0.796 $\pm$ 0.052 & 0.508 $\pm$ 0.069 & 15.41 $\pm$ 5.100
& 124.0 $\pm$ 20.60 & 0.761 $\pm$ 0.037 & 0.485 $\pm$ 0.089 & 19.46 $\pm$ 6.577 \\

Flow Matching\cite{lipman2022flow}
& 106.3 $\pm$ 29.38 & 0.933 $\pm$ 0.035 & 0.769 $\pm$ 0.089 & 4.530 $\pm$ 0.878
& 89.72 $\pm$ 27.09 & 0.901 $\pm$ 0.060 & 0.623 $\pm$ 0.088 & 9.792 $\pm$ 3.041
& 93.26 $\pm$ 11.58 & 0.842 $\pm$ 0.019 & 0.579 $\pm$ 0.077 & 16.56 $\pm$ 4.130 \\

\rowcolor{gray!15}
\textbf{GDM}
& \textbf{72.13 $\pm$ 17.37} & \textbf{0.959 $\pm$ 0.023} & \textbf{0.828 $\pm$ 0.077} & \textbf{3.640 $\pm$ 0.834}
& \textbf{65.69 $\pm$ 24.00} & \textbf{0.926 $\pm$ 0.053} & \textbf{0.687 $\pm$ 0.088} & \textbf{8.115 $\pm$ 2.957}
& \textbf{67.28 $\pm$ 14.87} & \textbf{0.888 $\pm$ 0.032} & \textbf{0.634 $\pm$ 0.077} & \textbf{12.15 $\pm$ 4.684} \\
\bottomrule
\end{tabular}%
}
\end{table*}

\begin{figure}[!t]
\centering
\includegraphics[width=0.985\linewidth]{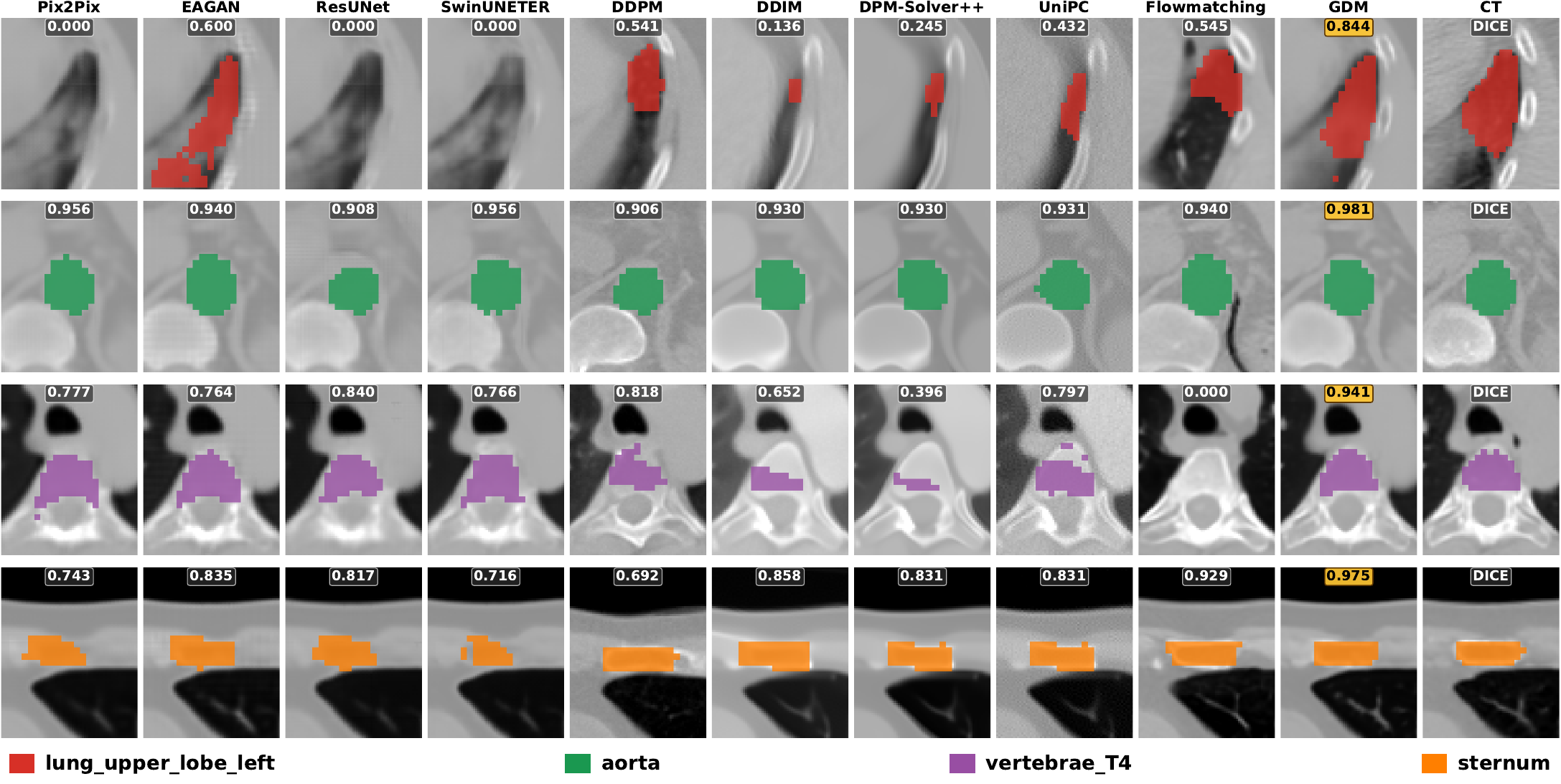}
\caption{Segmentation on synthesized CT volumes from the SynthRAD2025 dataset using TotalSegmentator\cite{wasserthal2023totalsegmentator}. The displayed structures are the left upper lung lobe, liver, aorta, T4 vertebra, and sternum.}
\label{fig:seg_results}
\end{figure}

\begin{figure}[t]
\centering
\includegraphics[width=0.985\linewidth]{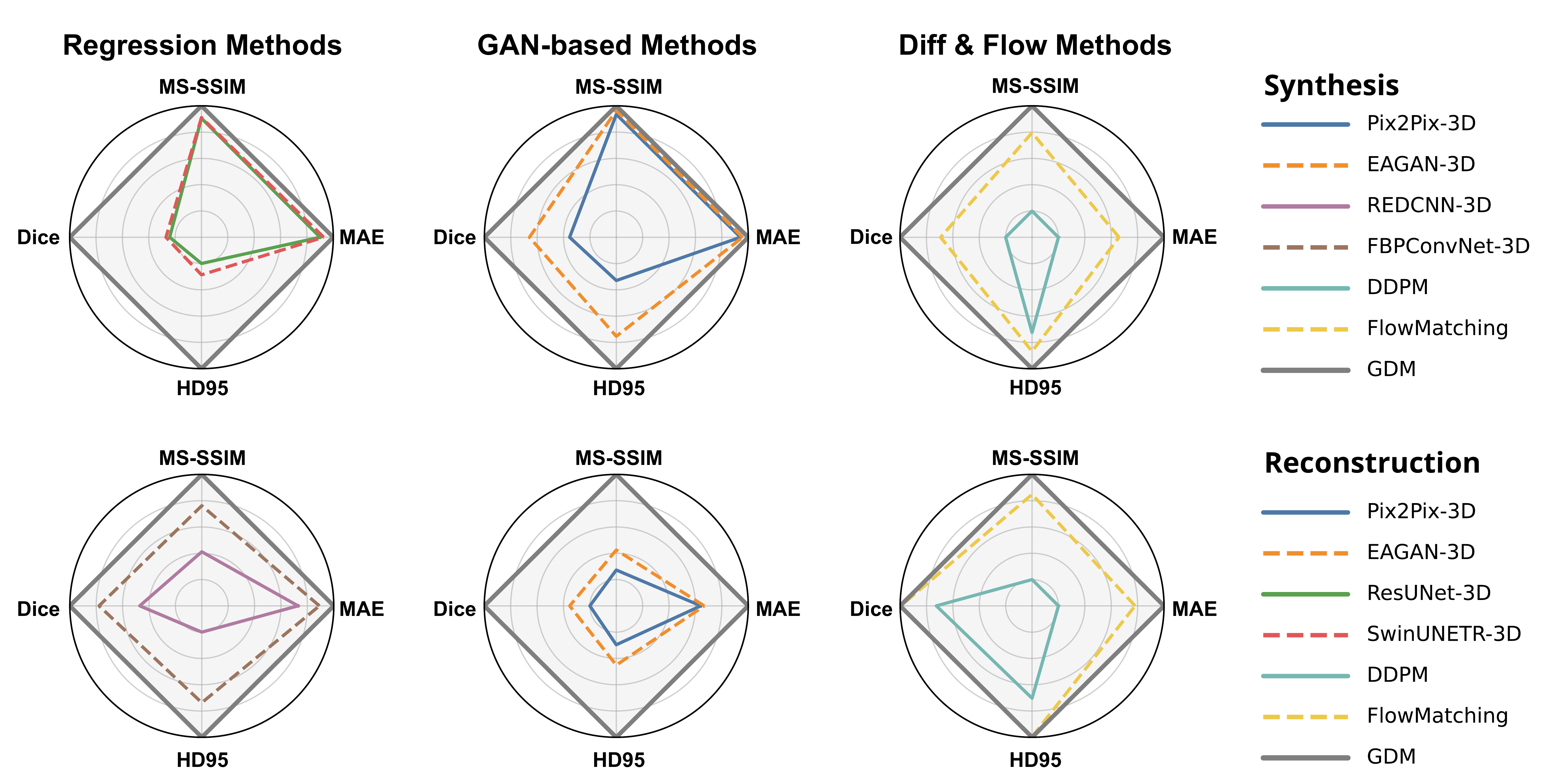}
\caption{Radar plots of normalized performance on MRI-to-CT synthesis and SVCT reconstruction. Metrics were jointly min--max normalized across methods for each task; MAE and HD95 were reversed so that higher values indicate better performance. Inner and outer rings correspond to the minimum and maximum normalized values, respectively.}
\label{fig:radar_main}
\end{figure}

\section{Experimental Setup and Results}

\begin{figure*}[t]
\centering
\includegraphics[width=0.985\linewidth]{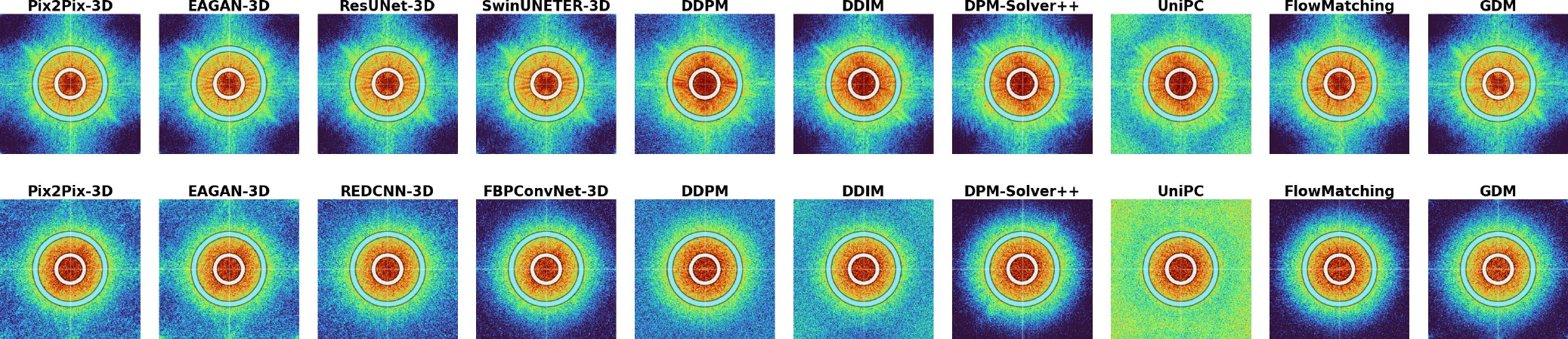}
\caption{Residual noise power spectrum (NPS) analysis\cite{kijewski1987noise}. The top row corresponds to MRI-to-CT synthesis and the bottom row to SVCT reconstruction. The inner and outer circles denote $0.2\times$ and $0.5\times$ Nyquist frequencies, respectively. GDM shows lower residual energy in both low- and high-frequency bands, indicating better global consistency and finer detail preservation.}
\label{fig:nps}
\end{figure*}

\begin{figure*}[t]
\centering
\includegraphics[width=0.985\linewidth]{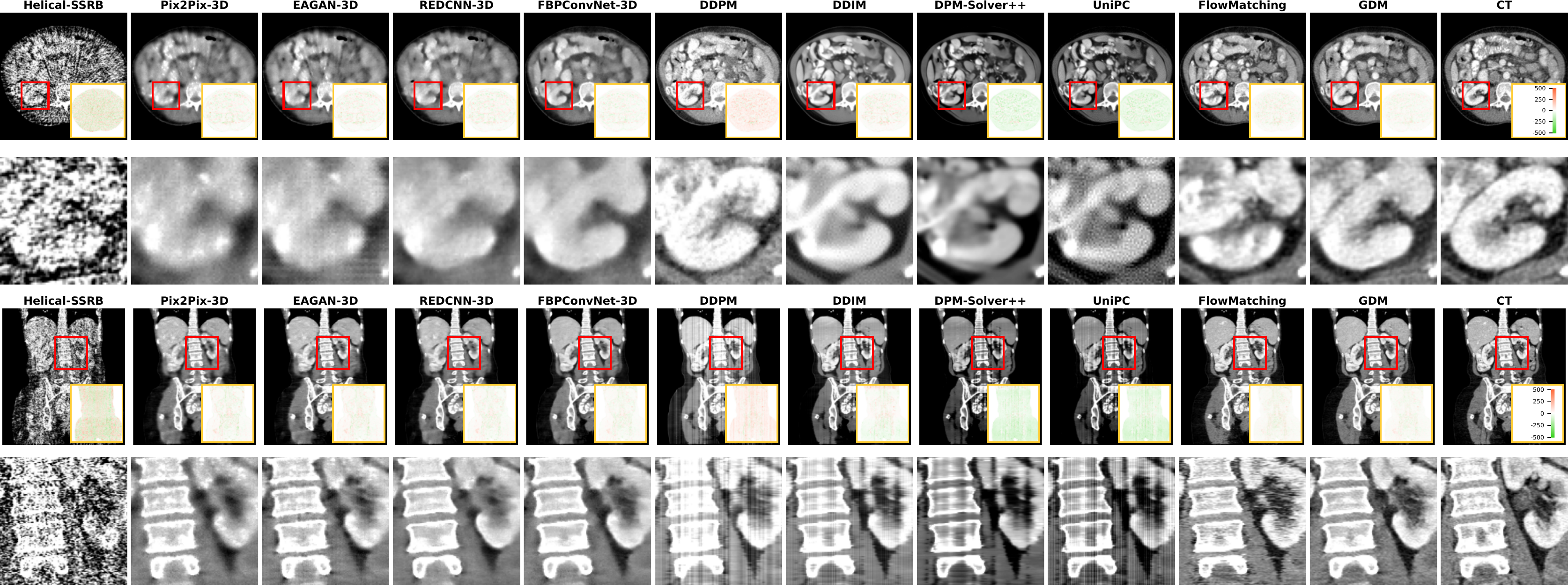}
\caption{SVCT reconstruction results from 60 views. Rows 1--2 show axial views (WW/WL = 350/40 HU), and rows 3--4 show coronal views (WW/WL = 350/40 HU). Rows 2 and 4 show enlarged ROIs marked by the red boxes. Insets are signed error maps relative to the reference CT.}
\label{fig:svct_vis}
\end{figure*}

\begin{figure}[t]
\centering
\includegraphics[width=0.985\linewidth]{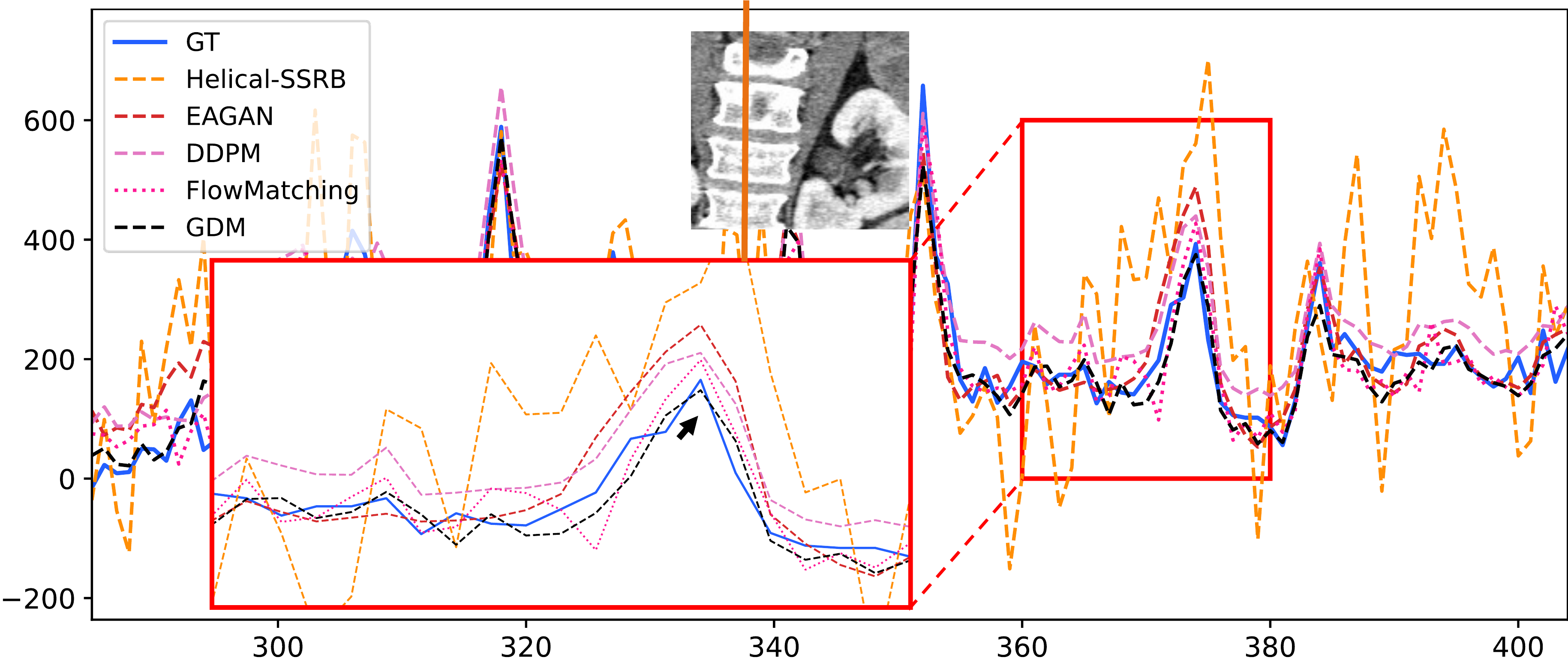}
\caption{Intensity profiles along the indicated line in the coronal ROI of SVCT reconstruction. The red box marks the enlarged region. }
\label{fig:profile}
\end{figure}

\subsection{Datasets and Implementation Details}

\subsubsection{Datasets}
We used two multi-center clinical datasets. For MRI-to-CT synthesis, we used 513 paired MRI--CT cases from SynthRAD2025~\cite{thummerer2025synthrad2025}, covering head-and-neck, thoracic, and abdominal scanning, using organizer-provided registrations. For SVCT reconstruction, we used 120 cases from the TCIA helical projection data collection~\cite{moen2021lowdose}. Header-derived scan geometry was vendor-specific for Siemens and GE scanners. Helical projections were converted to fan-beam sinograms using single-slice rebinning (SSRB)~\cite{noo1999single}, resulting in Siemens ($n_u{=}736$, $d_u{=}1.2858$ mm, $\mathrm{DSO}{=}595.0$ mm, $\mathrm{DDO}{=}490.6$ mm) and GE ($n_u{=}888$, $d_u{=}1.0239$ mm, $\mathrm{DSO}{=}538.52$ mm, $\mathrm{DDO}{=}408.226$ mm) fan-beam geometry. Full-view sinograms were reconstructed by Hann-filtered FBP into $512\times512$ images with 0.9 mm voxel size as ground truth; sparse-view inputs were obtained by uniformly downsampling the rebinned sinograms to 60 views and reconstructing them identically. Both datasets were randomly split 70\%/10\%/20\% for training/validation/testing.
\subsubsection{Implementation Details}
We compared GDM with regression, GAN, diffusion, and flow baselines. Shared baselines for both tasks were Pix2Pix-3D\cite{isola2017image}, EAGAN-3D\cite{yu2019ea}, DDPM\cite{ho2020denoising}, DDIM\cite{song2020denoising}, DPM-Solver++\cite{lu2025dpm}, UniPC\cite{zhao2023unipc}, and Flow Matching\cite{lipman2022flow}. ResUNet-3D\cite{xiao2018weighted} and SwinUNETR-3D\cite{hatamizadeh2021swin} were additionally used for MRI-to-CT synthesis, while helical SSRB\cite{noo1999single}, REDCNN-3D\cite{chen2017low}, and FBPConvNet-3D\cite{jin2017deep} were added for SVCT reconstruction. Except for diffusion/flow methods, all baselines were implemented in 3D.
GDM and all trainable non-diffusion baselines were optimized with AdamW for 1000 epochs using an initial learning rate of $10^{-3}$, linear warm-up, cosine decay, and patch size $32\times128\times128$. Diffusion/flow baselines were trained slice-wise for 100k steps with learning rate $2\times10^{-4}$. At inference, non-diffusion methods used sliding-window inference with window size $32\times128\times128$ and 70\% overlap. DDPM used 1000 sampling steps, whereas DDIM, DPM-Solver++, UniPC, and Flow Matching used 100. Evaluation metrics were mean absolute error (MAE), multi-scale structural similarity (MS-SSIM), Dice coefficient (Dice), and 95th-percentile Hausdorff distance (HD95). Segmentation used TotalSegmentator\cite{wasserthal2023totalsegmentator}.

For GDM, MedNext-3D~\cite{roy2023mednext} was used as the generator. MedVAE-3D\cite{varma2025medvae} with $4\times$ compression was fine-tuned on each training set and then frozen as the drifting feature extractor. For drift computation, 12 sub-volumes of size $16\times32\times32$ were randomly sampled from each input patch. Training used four NVIDIA H100 GPUs with distributed data parallelism and batch size 8 per GPU, yielding 384 interacting sub-volumes per iteration. Temperatures were fixed to $\tau\in\{0.004,0.01,0.04\}$, and the drifting loss in Eq.~(\ref{eq:mgda_objectives}) was pre-scaled by $\lambda=3\times10^{-4}$ before MGDA.

\subsection{MRI-to-CT Synthesis Results}

Fig.~\ref{fig:mri_ct_vis} shows MRI-to-CT synthesis results on SynthRAD2025. CNN/GAN baselines recover coarse anatomy but over-smooth thin bone structures and tissue interfaces. Diffusion- and flow-based baselines often produce sharper, more realistic outputs, yet the ROIs and signed error maps still reveal structural mismatch, anatomical deviation, and intensity bias from the patient-specific target. Because full 3D diffusion is computationally prohibitive, these methods are implemented slice-wise, causing weaker volumetric consistency and coronal discontinuities. In contrast, GDM is visually closest to the reference CT, with sharper local structures, cleaner density transitions, more localized residuals, and more coherent downstream segmentations in Fig.~\ref{fig:seg_results}.

Table~\ref{tab:mri_ct_results} confirms these findings. GDM ranks first on all four metrics, improving intensity accuracy and structural quality over regression and GAN baselines while better preserving patient-specific anatomy than diffusion and flow baselines without sacrificing realistic CT appearance. This reflects the central trade-off in deterministic medical generation: direct mappings tend to over-smooth, whereas stronger generative modeling may improve realism at the cost of  fidelity. Furthermore, the top row of Fig.~\ref{fig:nps} shows lower residual energy for GDM in both low- and high-frequency, indicating better global consistency and finer detail preservation. The upper radar plot in Fig.~\ref{fig:radar_main} shows the same trend, with the most balanced overall profile among all four metrics.

\begin{table}[!t]
\centering
\caption{Quantitative evaluation results on SVCT reconstruction. Best results are highlighted in bold.}
\label{tab:svct_results}
\renewcommand{\arraystretch}{1.3}
\setlength{\tabcolsep}{4pt}
\resizebox{\columnwidth}{!}{
\begin{tabular}{lcccc}
\toprule
\rowcolor{sectionblue}
\multicolumn{5}{c}{\textbf{SVCT Reconstruction Results on TCIA Dataset}\cite{moen2021lowdose}} \\
\midrule
\textbf{Method} & \textbf{MAE}$\downarrow$ & \textbf{MS-SSIM}$\uparrow$ & \textbf{Dice}$\uparrow$ & \textbf{HD95}$\downarrow$ \\
\midrule
Helical SSRB\cite{noo1999single}    & 114.0 $\pm$ 3.370 & 0.860 $\pm$ 0.008 & 0.830 $\pm$ 0.079 & 6.647 $\pm$ 4.203 \\
Pix2Pix-3D\cite{isola2017image}     & 36.86 $\pm$ 2.190 & 0.975 $\pm$ 0.003 & 0.874 $\pm$ 0.063 & 5.139 $\pm$ 3.311 \\
EAGAN-3D\cite{yu2019ea}             & 36.38 $\pm$ 2.320 & 0.977 $\pm$ 0.003 & 0.885 $\pm$ 0.064 & 4.571 $\pm$ 3.390 \\
REDCNN-3D\cite{chen2017low}         & 35.20 $\pm$ 2.920 & 0.977 $\pm$ 0.004 & 0.894 $\pm$ 0.048 & 5.490 $\pm$ 3.744 \\
FBPConvNet-3D\cite{jin2017deep}     & 32.17 $\pm$ 1.700 & 0.982 $\pm$ 0.002 & 0.917 $\pm$ 0.026 & 3.526 $\pm$ 2.094 \\
DDIM\cite{song2020denoising}        & 65.29 $\pm$ 27.86 & 0.966 $\pm$ 0.019 & 0.923 $\pm$ 0.017 & 3.065 $\pm$ 1.449 \\
DPM-Solver++\cite{lu2025dpm}        & 104.8 $\pm$ 42.37 & 0.921 $\pm$ 0.064 & 0.917 $\pm$ 0.016 & 3.294 $\pm$ 1.615 \\
UniPC\cite{zhao2023unipc}           & 104.7 $\pm$ 40.96 & 0.924 $\pm$ 0.059 & 0.907 $\pm$ 0.034 & 4.091 $\pm$ 3.105 \\
DDPM\cite{ho2020denoising}          & 45.10 $\pm$ 5.430 & 0.974 $\pm$ 0.002 & 0.913 $\pm$ 0.026 & 3.650 $\pm$ 2.508 \\
Flow Matching\cite{lipman2022flow}  & 34.24 $\pm$ 1.430 & 0.984 $\pm$ 0.002 & 0.933 $\pm$ 0.013 & 2.580 $\pm$ 0.090 \\
\rowcolor{gray!15}
\textbf{GDM} & \textbf{30.14 $\pm$ 1.180} & \textbf{0.986 $\pm$ 0.001} & \textbf{0.934 $\pm$ 0.021} & \textbf{2.563 $\pm$ 1.125} \\
\bottomrule
\end{tabular}
}
\end{table}

\subsection{SVCT Reconstruction Results}

Fig.~\ref{fig:svct_vis} shows 60-view CT reconstruction results on TCIA. Helical SSRB suffers from severe streak artifacts and noise contamination. CNN/GAN baselines suppress most large-scale artifacts, but fine structures and soft-tissue boundaries remain over-smoothed. Diffusion- and flow-based baselines often appear sharper and more realistic, yet closer inspection of the ROIs reveals evident mismatch with the reference CT, including distorted kidney shape, inaccurate boundaries, and inconsistent internal appearance. In contrast, GDM remains closest to the reference CT, with better preservation of thin structures and more faithful renal anatomy. This is further supported by the intensity profiles in Fig.~\ref{fig:profile}, where GDM follows the reference CT more closely and better preserves local intensity transitions, especially near sharp boundaries. Consistent with these visual and profile-based observations, Table~\ref{tab:svct_results} shows that GDM again ranks first on all four metrics.

The lower radar plot in Fig.~\ref{fig:radar_main} further summarizes this trend. GDM shows the most balanced profile across regression, GAN-based, and diffusion/flow-based methods. In particular, it avoids the common SVCT trade-off between artifact suppression and structural recovery, as well as the tendency of sharper-looking methods to sacrifice anatomical faithfulness. The bottom row of Fig.~\ref{fig:nps} provides complementary frequency-domain evidence: GDM exhibits lower residual energy in both low- and high-frequency bands, indicating better global consistency and finer detail preservation. In SVCT, this suggests more effective suppression of large-scale streaking artifacts while preserving boundaries and thin structures.

\begin{figure}[t]
\centering
\includegraphics[width=0.985\linewidth]{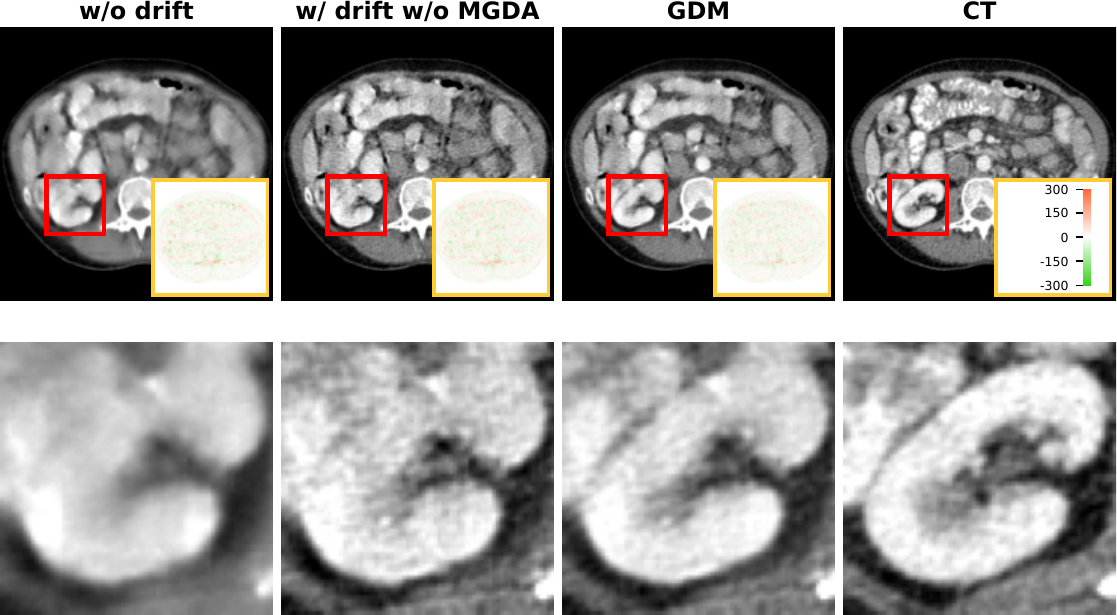}
\caption{Ablation of drifting and MGDA on SVCT reconstruction. Without drifting, the reconstruction becomes over-smoothed. Adding drifting without MGDA recovers sharper details but introduces local distortions.}
\label{fig:ablation_drifting_mgda}
\end{figure}
\subsection{Ablation Studies}

\begin{table}[!t]
\centering
\caption{Comparison of inference cost. Inference time is measured per test volume ($512^{3}$) on a single NVIDIA H100 GPU.}
\label{tab:cost_results}
\renewcommand{\arraystretch}{1.2}
\setlength{\tabcolsep}{4pt}
\resizebox{\columnwidth}{!}{
\begin{tabular}{llcc}
\toprule
\textbf{Method} & \textbf{Paradigm} & \textbf{Sampling Steps} & \textbf{Time / Volume} \\
\midrule
Pix2Pix-3D\cite{isola2017image}    & GAN         & 1    & 9.69 s \\
EAGAN-3D\cite{yu2019ea}            & GAN         & 1    & 10.12 s \\
ResUNet-3D\cite{xiao2018weighted}  & CNN         & 1    & 10.32 s \\
SwinUNETR-3D\cite{hatamizadeh2021swin} & Transformer & 1    & 1 min 28 s \\
FBPConvNet-3D\cite{jin2017deep}    & CNN         & 1    & 35.63 s \\
REDCNN-3D\cite{chen2017low}        & CNN         & 1    & 2 min 13 s \\
Flow Matching\cite{lipman2022flow} & Flow        & 100  & 31 min 20 s \\
DDIM\cite{song2020denoising}       & Diffusion   & 100  & 48 min 45 s \\
DPM-Solver++\cite{lu2025dpm}       & Diffusion   & 100  & 51 min 41 s \\
UniPC\cite{zhao2023unipc}          & Diffusion   & 100  & 43 min 33 s \\
DDPM\cite{ho2020denoising}         & Diffusion   & 1000 & 8 h 6 min 26 s \\
\rowcolor{gray!15}
\textbf{GDM} & \textbf{Drifting} & \textbf{1} & \textbf{1 min 33 s} \\
\bottomrule
\end{tabular}
}
\end{table}

\begin{table}[!t]
\centering
\caption{Ablation study of the drifting mechanism and MGDA on MRI-to-CT synthesis and SVCT reconstruction.}
\label{tab:ablation_results}
\renewcommand{\arraystretch}{1.3}
\setlength{\tabcolsep}{4pt}
\resizebox{\columnwidth}{!}{
\begin{tabular}{lcccc}
\toprule
\rowcolor{sectionblue}
\multicolumn{5}{c}{\textbf{MRI-to-CT Synthesis Results on SynthRAD2025 Dataset}} \\
\midrule
\textbf{Method} & \textbf{MAE}$\downarrow$ & \textbf{MS-SSIM}$\uparrow$ & \textbf{Dice}$\uparrow$ & \textbf{HD95}$\downarrow$ \\
\midrule
w/o drifting             & 70.21 (ref.) & 0.919 (ref.) & 0.650 (ref.) & 14.095 (ref.) \\
w/ drifting w/o MGDA     & 71.83 (\degrade{+1.62}) & 0.918 (0.000) & 0.724 (\improve{+0.074}) & 7.405 (\improve{-6.690}) \\
\rowcolor{gray!15}
GDM                      & 68.37 (\improve{-1.84}) & 0.924 (\improve{+0.006}) & 0.716 (\improve{+0.067}) & 7.969 (\improve{-6.126}) \\
\specialrule{\heavyrulewidth}{0pt}{0pt}
\rowcolor{sectionblue}
\multicolumn{5}{c}{\textbf{SVCT Reconstruction Results on TCIA Dataset}} \\
\midrule
\textbf{Method} & \textbf{MAE}$\downarrow$ & \textbf{MS-SSIM}$\uparrow$ & \textbf{Dice}$\uparrow$ & \textbf{HD95}$\downarrow$ \\
\midrule
w/o drifting             & 36.69 (ref.) & 0.978 (ref.) & 0.914 (ref.) & 3.733 (ref.) \\
w/ drifting w/o MGDA     & 42.46 (\degrade{+5.77}) & 0.973 (\degrade{-0.005}) & 0.896 (\degrade{-0.018}) & 4.695 (\degrade{+0.962}) \\
\rowcolor{gray!15}
GDM                      & 30.14 (\improve{-6.55}) & 0.986 (\improve{+0.008}) & 0.934 (\improve{+0.020}) & 2.563 (\improve{-1.170}) \\
\bottomrule
\end{tabular}
}
\end{table}
\subsubsection{Computational Cost Analysis}
To examine whether GDM retains practical efficiency, Table~\ref{tab:cost_results} compares model complexity and inference cost. GDM performs inference with a single forward pass, whereas Flow Matching, DDIM, DPM-Solver++, and UniPC require 100 sampling steps and DDPM requires 1000. Noticeably, GDM reconstructs a $512^3$ volume in 1 min 33 s, which is substantially faster than these multi-step inference baselines and remains in the same order of magnitude as CNN- and GAN-based one-step models.
\begin{figure}[t]
\centering
\includegraphics[width=0.985\linewidth]{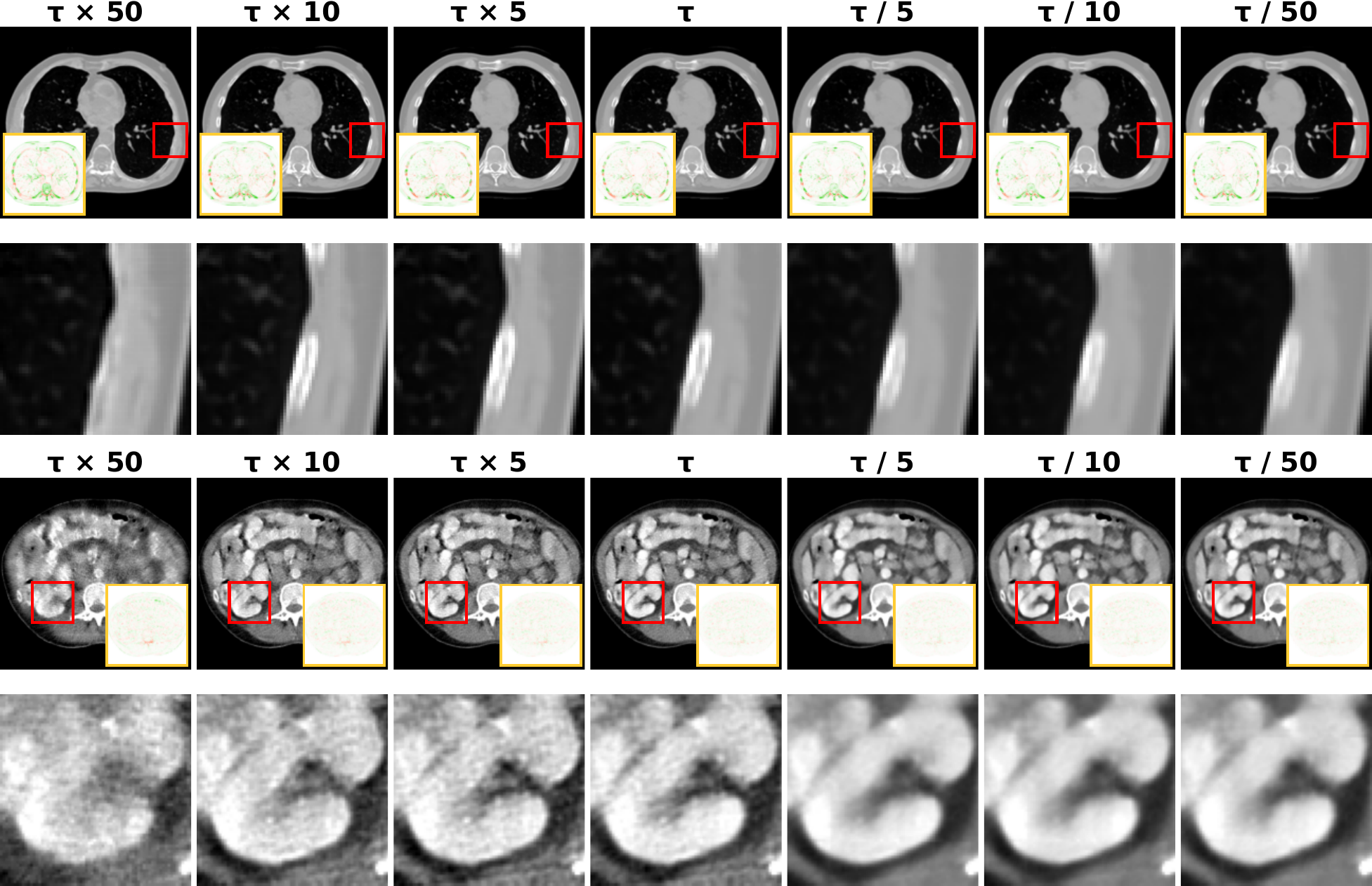}
\caption{Ablation of drifting temperature $\tau\in\{0.004,0.01,0.04\}$ on synthesis (top) and reconstruction (bottom). Larger $\tau$ yields stronger generative effects and sharper appearance but may increase structural instability, whereas smaller $\tau$ produces over-smoothed outputs.}
\label{fig:ablation_tau}
\end{figure}

\begin{table}[!t]
\centering
\caption{Ablation of drifting temperature $\tau$. Parenthetical values denote differences from GDM.}
\label{tab:tau_ablation_results}
\renewcommand{\arraystretch}{1.2}
\setlength{\tabcolsep}{4pt}
\resizebox{\columnwidth}{!}{
\begin{tabular}{ccccc}
\toprule
\multirow{2}{*}{\textbf{$\tau$}}
& \multicolumn{2}{>{\columncolor{sectionblue}}c}{\textbf{MRI-to-CT Synthesis}}
& \multicolumn{2}{>{\columncolor{sectionblue}}c}{\textbf{SVCT Reconstruction}} \\
\cmidrule(lr){2-3} \cmidrule(lr){4-5}
& \textbf{MAE}$\downarrow$ & \textbf{Dice}$\uparrow$
& \textbf{MAE}$\downarrow$ & \textbf{Dice}$\uparrow$ \\
\midrule
$\tau \times 50$ & 102.46 (\degrade{+34.09}) & 0.531 (\degrade{-0.185}) & 40.96 (\degrade{+10.82}) & 0.908 (\degrade{-0.026}) \\
$\tau \times 10$ & 79.41 (\degrade{+11.05})  & 0.711 (\degrade{-0.005}) & 34.67 (\degrade{+4.53})  & 0.931 (\degrade{-0.003}) \\
$\tau \times 5$  & 75.20 (\degrade{+6.83})   & 0.723 (\improve{+0.007}) & 32.82 (\degrade{+2.68})  & 0.934 (0.000) \\
\rowcolor{gray!15}
GDM              & 68.37 (ref.)              & 0.716 (ref.)             & 30.14 (ref.)             & 0.934 (ref.) \\
$\tau / 5$       & 71.24 (\degrade{+2.87})   & 0.671 (\degrade{-0.046}) & 28.48 (\improve{-1.66})  & 0.934 (0.000) \\
$\tau / 10$      & 68.24 (\improve{-0.13})   & 0.673 (\degrade{-0.044}) & 28.67 (\improve{-1.47})  & 0.932 (\degrade{-0.002}) \\
$\tau / 50$      & 69.80 (\degrade{+1.44})   & 0.668 (\degrade{-0.049}) & 28.79 (\improve{-1.35})  & 0.933 (\degrade{-0.001}) \\
\bottomrule
\end{tabular}
}
\end{table}

\subsubsection{Ablation Study of Drifting and MGDA}

Table~\ref{tab:ablation_results} and Fig.~\ref{fig:ablation_drifting_mgda} show a clear progression from \emph{w/o drift} to \emph{w/ drift w/o MGDA} to the full GDM. In the SVCT example of Fig.~\ref{fig:ablation_drifting_mgda}, omitting drifting causes noticeable over-smoothing, blurred boundaries, and missing local structures in the zoomed region. Although MAE and MS-SSIM change only modestly, the visual evidence and the larger drops in Dice and HD95 indicate substantially weakened anatomical recognition, confirming drifting as the key mechanism that injects target-domain plausibility into the generator.

However, drifting alone is insufficient. Relative to \emph{w/o drift}, the \emph{w/ drift w/o MGDA} variant alleviates over-smoothing and appears sharper and more realistic, but also introduces local distortions, showing that drifting without explicit coordination with fidelity supervision can yield visually plausible yet structurally inaccurate results. Table~\ref{tab:ablation_results} shows the same trend: for MRI-to-CT synthesis, removing MGDA slightly improves Dice and HD95, but worsens MAE and does not improve MS-SSIM; for SVCT reconstruction, it degrades all four metrics. Thus, drifting and MGDA play complementary roles: drifting prevents collapse to an over-smoothed average solution, while MGDA balances fidelity and distribution alignment to steer optimization toward a more stable Pareto solution.

\subsubsection{Ablation Study of Different $\tau$}

To study the sensitivity of drifting to affinity scaling, we further analyze the effect of different temperature settings in the drifting field. Specifically, we scale the default setting $\tau\in\{0.004,0.01,0.04\}$ by factors of $5$, $10$, and $50$ and by their reciprocals. Table~\ref{tab:tau_ablation_results} and Fig.~\ref{fig:ablation_tau} show a consistent trend across both tasks: increasing $\tau$ strengthens the generative effect and yields sharper appearances, but excessively large values destabilize local structures, whereas decreasing $\tau$ suppresses the generative effect and leads to more conservative, over-smoothed outputs. 

\section{Discussion and Conclusion}

GDM should be viewed not only as a method for MRI-to-CT synthesis and SVCT reconstruction, but also as a general framework for paired conditional medical image generation with deterministic input-target correspondence. By decoupling one-step image generation from training-time distribution alignment and coordinating them through multi-objective optimization, GDM can be readily extended to other paired translation, reconstruction, and restoration tasks that require fidelity, plausibility, and efficiency.

Our results further suggest that GDM is relatively robust under complex target distributions and imperfect pairwise correspondence. This is particularly relevant to MRI-to-CT synthesis, where multi-center, multi-scanner, and multi-body-region data induce a highly heterogeneous CT distribution, while MRI--CT registration errors introduce residual mismatch in the paired supervision. In such settings, diffusion models may improve realism but remain unstable in recovering the paired patient-specific target. By retaining direct fidelity supervision as an anatomical anchor and introducing drifting-based distribution alignment in structured medical feature spaces, GDM produces more stable and anatomically grounded results.

In this work, we presented GDM, a one-step generative drifting framework for conditional medical image generation. By combining multi-level feature-based drifting with MGDA-based gradient coordination in the shared output space, GDM jointly promotes patient-specific fidelity and distribution-level plausibility while preserving efficient inference. Experiments on MRI-to-CT synthesis and SVCT reconstruction demonstrate consistent advantages over regression, GAN-based, and diffusion/flow-based baselines. More broadly, GDM offers a practical and extensible framework for deterministic conditional medical image generation, where realism, fidelity, and efficiency must be optimized together.

\section*{Acknowledgment}
\label{acknowledgment}
The authors gratefully acknowledge the scientific support and HPC resources provided by the Erlangen National High Performance Computing Center (NHR@FAU) of the Friedrich-Alexander-Universität Erlangen-Nürnberg (FAU). The hardware is partially funded by the German Research Foundation (DFG).
\bibliographystyle{IEEEtran}
\bibliography{reference}

\end{document}